\pdfoutput=1
\documentclass{article}

\usepackage[letterpaper,margin=1in]{geometry}
\usepackage{times}
\usepackage[numbers,sort&compress]{natbib}

\usepackage[T1]{fontenc}

\usepackage[utf8]{inputenc}
\usepackage{microtype}
\usepackage{amsmath,amssymb}
\usepackage{booktabs}
\usepackage{array}
\usepackage{subcaption}
\newcolumntype{L}[1]{>{\raggedright\arraybackslash}p{#1}}
\usepackage{graphicx}
\usepackage[table]{xcolor}
\definecolor{ourrow}{HTML}{E6EFF8}
\usepackage{tikz}
\definecolor{boxedge}{HTML}{1F3B5C}
\definecolor{stA}{HTML}{0072B2}\definecolor{stAbg}{HTML}{EEF5FB}
\definecolor{stB}{HTML}{00785A}\definecolor{stBbg}{HTML}{ECF7F2}
\definecolor{stC}{HTML}{B34700}
\definecolor{stD}{HTML}{8E4A79}\definecolor{stDbg}{HTML}{FAF0F6}
\newcommand{\stagebadge}[2]{\tikz[baseline=(n.base)]\node[circle,fill=#1,text=white,font=\scriptsize\bfseries,inner sep=0pt,minimum size=11pt](n){#2};}
\usetikzlibrary{arrows.meta,positioning}
\usepackage{enumitem}
\usepackage{algorithm}
\usepackage{algpseudocode}
\algtext*{EndWhile}\algtext*{EndFor}\algtext*{EndIf}\algtext*{EndFunction}\algtext*{EndProcedure}
\usepackage{hyperref}
\hypersetup{colorlinks=true,linkcolor=blue!60!black,citecolor=blue!60!black,urlcolor=blue!60!black}
\usepackage{cleveref}

\newcommand{\ck}[1]{\textsf{#1}}

\title{\texttt{ThinkBit}: Search Codewords, Recalibrate Scales,\\ Select by Reasoning}

\author{Rhea Sanjay Sukthanker \qquad Pashmina Cameron\\
Microsoft Applied Sciences Group\\[6pt]
\texttt{rsukthanker@microsoft.com} \qquad
\texttt{pcameron@microsoft.com}}
\date{}

\usepackage{float}
\usepackage[section]{placeins}
\begin{document}
\raggedbottom
\maketitle

\begin{abstract}
Reasoning language models generate thousands of tokens per answer, and generation speed is limited by memory bandwidth, so storing weights in two bits instead of \texttt{bfloat16} can potentially make inference 8$\times$ cheaper. Two-bit methods are usually judged by perplexity and multiple-choice question-answering accuracy, both of which say little about whether a model can still reason through more difficult problems. We present \texttt{ThinkBit}, a strictly scalar two-bit framework built to preserve reasoning, in which every weight is an integer codeword times a \texttt{bfloat16} scale shared by a group of weights. We first use a sensitivity metric to assign each layer a ternary, 2-bit, or 3-bit tier to allocate bitwidths optimally among layers. This is followed by a layerwise Gumbel-softmax-based scalar quantization algorithm that searches for optimal codewords block by block. Since a layerwise scheme does not capture how quantized layers interact with each other, we recalibrate scales, with the codewords frozen, using a knowledge-distillation-based loss against the teacher model. We then perform a short quantization-aware training (QAT) stage that lets the model mimic the base model's distribution by reassigning codewords. Finally, a paired reasoning test on training problems selects the final checkpoint between the recalibration stage and the QAT stage. On Qwen3-1.7B at matched size, our recalibrated checkpoint scores 46.7 on MATH-500 against 30.0 for EfficientQAT trained on the same data, and 73.0 after QAT on reasoning traces; at 0.52\,GB it is smaller than every published 2-bit GGUF of the model yet more accurate on MATH-500. We show that this framework scales and up to 14B, a 3.74\,GB model scores 94.8 on MATH-500, against 97.5 for the 29.5\,GB \texttt{bfloat16} model.
\end{abstract}

\section{Introduction}
Reasoning language models answer with long chains of thought, often tens of thousands of tokens long. Each generated token requires reading all of the model's weights from memory, so generation speed is capped by memory bandwidth rather than compute. Storing weights in two bits instead of \texttt{bf16} reduces weight memory and the memory traffic per token by up to $8\times$. Above three bits per weight, scalar quantization formats, which store one integer per weight and one scale per group of weights, are already close to lossless. At two bits, however, the strongest methods go beyond this format. They use vector codebooks, rotations, zero-points, or distillation-based quantization-aware training (QAT) with small groups~\citep{egiazarian2024aqlm,malinovskii2024pvtuning,du2024bitdistiller,liu2025paretoq,chen2024efficientqat}. Scalar formats remain the easiest to deploy, because dequantizing a weight takes a single multiplication. However, it is still an open question whether a two-bit scalar format can preserve a model's ability to reason through difficult problems.

\begin{figure}[!htbp]
\centering
\begin{subfigure}{\linewidth}
\centering
\includegraphics[width=\linewidth]{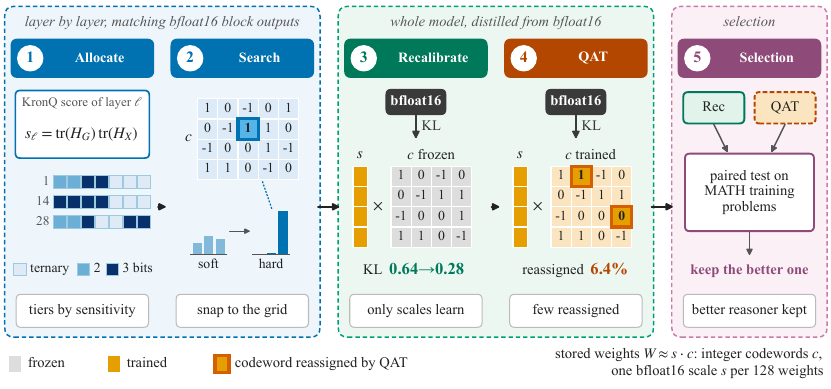}
\caption{The proposed framework (\texttt{ThinkBit}), has five stages. (1)~Kronecker-factored Hessian sensitivity (KronQ \citep{kronq}) assigns each layer a ternary, 2-bit, or 3-bit tier. (2)~GSQ \citep{dadgarnia2026gsq} starts from a layer-wise PTQ solution and searches for codewords. (3)~Scale recalibration freezes the codewords and trains only the stored bfloat16 (\texttt{bf16}) scales and the root-mean-square normalization (RMSNorm) gains. (4)~QAT reassigns codewords. (5)~A paired reasoning test on MATH training problems decides which checkpoint to keep.}
\label{fig:framework-a}
\end{subfigure}

\vspace{4pt}
\begin{subfigure}{\linewidth}
\centering
\includegraphics[width=\linewidth]{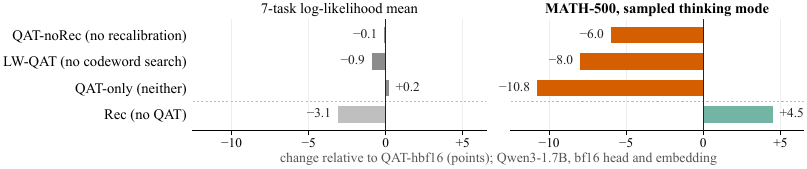}
\caption{Change in accuracy relative to the Qwen3-1.7B QAT checkpoint (\ck{QAT-hbf16}; names in \cref{tab:names}). All variants use the same \texttt{bf16} head and embedding. Removing codeword search, scale recalibration, or both keeps the 7-task mean in the range of 53.7--54.8, but lowers sampled MATH-500 accuracy from 43 to 35.0, 37.0, and 32.1 presented in \cref{tab:controls} (uniform 2-bit allocation instead of mixed tiers to 28.4). The recalibrated checkpoint without QAT (\ck{Rec-hbf16}) is \emph{lower} on the 7-task mean (51.5) but \emph{higher} on reasoning (47.5). Standard benchmarks therefore don't detect losses in reasoning correctly and could misguide checkpoint selection.}
\label{fig:framework-b}
\end{subfigure}
\caption{\textbf{Overview of our 2-bit quantization framework \texttt{ThinkBit} and importance of different stages}}
\label{fig:framework}
\end{figure}

Answering this question first requires fixing how two-bit models are evaluated. They are usually compared by perplexity and multiple-choice accuracy. These metrics barely separate configurations whose sampled reasoning accuracy differs by more than ten points (\cref{fig:framework}b), because likelihood based metrics, score single-token predictions and never check whether a long chain of thought eventually reaches a correct answer. A two-bit framework for reasoning models should therefore be both designed for and evaluated on reasoning tasks.

We design such a framework, \texttt{ThinkBit}, presented in \cref{fig:framework}a) with one primary constraint in mind. The constraint is that we want a strictly scalar quantization format. Every stored weight is an integer codeword multiplied by a \texttt{bf16} scale shared by a group of 128 weights, with no vector codebooks, rotations, or zero-points. Our proposed framework has five stages. First, sensitivity-based bit allocation assigns a precision to each layer. Second, block-wise codeword search chooses the initial codewords. Third, scale recalibration repairs the quantized model while leaving every codeword unchanged. Fourth, for even more flexibility, QAT reassigns the small number of codewords that still affect the output. Finally, a paired reasoning test on MATH training problems decides whether to keep the QAT checkpoint (\ck{QAT}) or the recalibrated checkpoint (\ck{Rec}). We compare both checkpoints against a \emph{QAT-only baseline} (\ck{QAT-only}), which skips the bit allocation, codeword search and recalibration phases.
\paragraph{What ``2-bit'' means.} The label ``2-bit'' does not by itself say how large a model is. Community quantizations portrayed as 2-bit usually keep the embeddings, the output head, and other sensitive tensors at higher precision, so their real storage cost is well above two bits per parameter. For example, Unsloth's 2-bit releases~\citep{unsloth2025qwen3gguf} in the GGUF file format of llama.cpp~\citep{gerganov2023llamacpp} store 2.82-3.62 bits per parameter for Qwen3-1.7B and 2.42-3.12 for Qwen3-14B. Similarly, PrismML's ternary Bonsai model~\citep{prismml2026bonsai}, whose weights need about $\log_2 3\approx1.585$ bits each, actually stores 2.15 bits per parameter for Qwen3-1.7B: every ternary weight sits in a 2-bit slot, with one FP16 scale per 128 weights. Hence, comparing such models by their nominal bit-width can hide differences in size of up to 50\%. We therefore compare models by \emph{whole-model size in bytes}, counting every stored tensor. By this measure, our 0.52\,GB Qwen3-1.7B checkpoint stores 2.42 bits per parameter including the output head and embeddings, which is smaller than every published 2-bit GGUF of the same model.

\paragraph{Our work makes the following key contributions:}
\begin{itemize}[leftmargin=1.2em,itemsep=2pt]
  \item \textbf{Two-bit models must be selected on reasoning tasks} On Qwen3-1.7B, removing the codeword search, scale recalibration, both, or the budget allocation stage keeps the 7-task QA task average at 53.7-54.8, within a $\pm$1.5-point margin. However, skipping these stages lowers MATH-500 accuracy from 43.0 after QAT stage to the range of 28.4--37.0 (\Cref{tab:controls}), showing that selecting two-bit models on reasoning tasks is crucial since that is where most of the degradations are seen. 
  \item \texttt{ThinkBit}\textbf{: A novel scalar two-bit framework built for reasoning.} We propose a novel five stage framework (\cref{fig:framework}a) (1) \texttt{Allocate}: Allocate precisions appropriately for each layer (2)  \texttt{Search}: Search the integer codewords block by block (3)  \texttt{Recalibrate}: Recalibrate the stored scales with the codewords frozen (4) \texttt{QAT}: Let quantization-aware training (QAT) reassign codewords and (5) \texttt{Selection}: Keep the better of the Stage-3 and Stage-4 checkpoints by a reasoning test. On Qwen3-1.7B our framework reaches 47.5 on sampled MATH-500 before QAT and 43.0 after, against 18.5--24.1 and 32.6--37.8 from tuned GPTQ initializations (\Cref{tab:reasoning}). %
  \item \textbf{Better two-bit reasoning per byte.} At matched size (about 0.5 GB), our recalibrated checkpoint (\ck{Rec-h2}) scores 46.7 on MATH-500, against 30.0 for EfficientQAT trained on the same data and tokens. At 0.52\,GB, smaller than every published 2-bit GGUF of the model, the selected checkpoint scores higher than all of them on MATH-500 (\Cref{tab:bytes}).%
\end{itemize}

\section{Methodology}
\label{sec:method}

\texttt{ThinkBit} turns a \texttt{bf16} model into a model whose linear-layer weights are stored as small integers. Stages~1--2 work layer by layer on calibration data; Stages~3--4 recover or train the whole model against the \texttt{bf16} model; Stage~5 only evaluates. \Cref{alg:full} in \cref{app:algorithm} gives the full procedure and \cref{tab:notation} in the appendix summarizes the notation. All models use the same experimental settings and we focus mainly on the Qwen3 \citep{yang2025qwen3} model family in our experiments.

\subsection{Weight format and bit accounting}
\label{sec:format}
Consider a linear layer $\ell$ with weight $W_\ell\in\mathbb{R}^{d_{\mathrm{out}}\times d_{\mathrm{in}}}$ and $n_\ell=d_{\mathrm{out}}d_{\mathrm{in}}$ weights. Each row is split into groups of $G=128$ consecutive input weights, and weight $(i,j)$ belongs to group $\gamma(j)=\lfloor j/G\rfloor$. The stored weight is an integer \emph{codeword} times one \texttt{bf16} \emph{scale} per group,
\begin{equation}
\widehat W_{ij}=s_{i,\gamma(j)}\,c_{ij},\qquad c_{ij}\in\mathcal{C}_b ,
\label{eq:format}
\end{equation}
where the grid $\mathcal{C}_b$ depends on the layer's \emph{tier} $b$: ternary $\{-1,0,1\}$, 2-bit $\{-2,-1,0,1\}$ or 3-bit $\{-4,\dots,3\}$. Dequantization is one multiplication; there is no zero-point, codebook or rotation. A tier costs
\begin{equation}
B_b=\log_2|\mathcal{C}_b|+\frac{16}{G}
\label{eq:bits}
\end{equation}
bits per weight: the codeword plus its share of the 16-bit scale. For example, ternary costs $\log_2 3+0.125=1.71$ bits, 2-bit $2.125$ and 3-bit $3.125$. The \emph{decoder bits per weight} (bpw) of a model averages $B_b$ over the linear layers of the transformer blocks, weighted by $n_\ell$; \emph{bits per parameter} also counts the embedding and output head and is the measure we use to compare models by size.

\subsection{\texttt{Allocate}: How do we allocate precision appropriately?}
\label{sec:stage1}
This stage of \texttt{ThinkBit} aims at optimally allocating precision to layers depending upon how sensitive they are to precision. Layers differ in how much rounding them hurts the model, so the budget should go where it matters. KronQ~\citep{kronq} estimates this from second-order information. Let $x_t$ and $y_t=W_\ell x_t$ be the input and output of layer $\ell$ at token position $t$ of a calibration sequence $\xi_1,\xi_2,\dots$ of tokens, and let $\mathcal{L}$ be the language-model loss. KronQ approximates the Hessian of $\mathcal{L}$ with respect to the layer's weights (vectorized column by column) by a Kronecker product of an input factor $H_{X,\ell}$ and an output factor $H_{G,\ell}$,
\begin{equation}
\begin{gathered}
\nabla^2_{\operatorname{vec}W_\ell}\mathcal{L}\;\approx\;H_{X,\ell}\otimes H_{G,\ell},\qquad H_{X,\ell}=\mathbb{E}_t\bigl[x_tx_t^{\top}\bigr],\qquad H_{G,\ell}=\mathbb{E}_t\bigl[g_tg_t^{\top}\bigr],\\
g_t=-\nabla_{y_t}\textstyle\sum_{t'}\log p_\theta(\tilde\xi_{t'+1}\mid \xi_{\le t'}),
\end{gathered}
\label{eq:kronq}
\end{equation}
where $p_\theta$ is the \texttt{bf16} model's next-token distribution and $g_t$ is the gradient of the loss at the layer's output. The next tokens $\tilde\xi_{t'+1}\sim p_\theta(\cdot\mid \xi_{\le t'})$ are sampled from the model itself, so $H_{X,\ell}\otimes H_{G,\ell}$ approximates the true Fisher information rather than the empirical Fisher tied to particular data labels; for the cross-entropy loss the true Fisher equals the Gauss--Newton approximation of the Hessian. The Kronecker form assumes, as in K-FAC~\citep{martens2015kfac}, that inputs and output gradients are independent and that gradients at different positions are uncorrelated; constant factors, such as the number of tokens, do not affect the allocation.

A rounding error $E_\ell=\widehat W_\ell-W_\ell$ changes the loss by a first-order term $\operatorname{vec}(E_\ell)^{\top}\nabla_{\operatorname{vec}W_\ell}\mathcal{L}$ and a second-order term. Because the labels are sampled from the model, the gradient has zero mean, $\mathbb{E}_{\tilde\xi}[g_t]=0$, so the first-order term vanishes in expectation and the loss rises, to second order, by
\begin{equation}
\Delta\mathcal{L}_\ell\approx\tfrac12\operatorname{vec}(E_\ell)^{\top}\bigl(H_{X,\ell}\otimes H_{G,\ell}\bigr)\operatorname{vec}(E_\ell)=\tfrac12\operatorname{tr}\bigl(H_{G,\ell}\,E_\ell\,H_{X,\ell}\,E_\ell^{\top}\bigr).
\label{eq:dloss}
\end{equation}
If the entries of $E_\ell$ are zero-mean and uncorrelated with variance $\sigma_{E,\ell}^2$, the expectation factorizes, $\mathbb{E}[\Delta\mathcal{L}_\ell]=\tfrac12\sigma_{E,\ell}^2\operatorname{tr}(H_{G,\ell})\operatorname{tr}(H_{X,\ell})$. The \emph{sensitivity} of layer $\ell$ is therefore
\begin{equation}
S_\ell=\operatorname{tr}(H_{G,\ell})\,\operatorname{tr}(H_{X,\ell})=\mathbb{E}_t\|g_t\|_2^2\cdot\mathbb{E}_t\|x_t\|_2^2 ,
\label{eq:sensitivity}
\end{equation}
which needs only the squared norms of inputs and output gradients, collected in one forward and backward pass per calibration sequence.

The error variance $\sigma_{E,\ell}^2$ depends on the tier. We summarize each tier by its \emph{rounding noise} $N_b$, the mean squared error when standard normal weights $\zeta_1,\dots,\zeta_G$ are rounded to $\mathcal{C}_b$ with the best scale for their group,
\begin{equation}
N_b=\mathbb{E}_{\zeta_j\sim\mathcal{N}(0,1)}\Bigl[\min_{s>0}\frac1G\sum_{j=1}^{G}\bigl(\zeta_j-s\,Q_b(\zeta_j/s)\bigr)^2\Bigr],
\label{eq:noise}
\end{equation}
where $Q_b$ rounds to the nearest point of $\mathcal{C}_b$. In practice we measure $N_b$ with the scale search of Stage~2 rather than the exact minimum, which gives slightly larger values (0.190, 0.150 and 0.041 against 0.187, 0.146 and 0.038 for ternary, 2-bit and 3-bit) and changes none of the conclusions below. Because the weights have unit variance, $N_b$ is the error relative to the weight variance, so $\sigma_{E,\ell}^2\approx N_b\operatorname{Var}(W_\ell)$; coarser grids have larger $N_b$. The expected loss of putting layer $\ell$ in tier $b$ is then $\tfrac12 S_\ell N_b\operatorname{Var}(W_\ell)$. Following KronQ~\citep{kronq}, we rank layers by $S_\ell N_b$, i.e.\ we drop the per-layer weight-variance factor and treat all layers as having equal weight variance. The allocation minimizes the total under a budget of $\bar B$ decoder bits per weight:
\begin{equation}
\min_{\{b_\ell\}}\ \sum_\ell S_\ell N_{b_\ell}\qquad\text{s.t.}\qquad \frac{\sum_\ell n_\ell B_{b_\ell}}{\sum_\ell n_\ell}\le\bar B .
\label{eq:alloc}
\end{equation}
This is a multiple-choice knapsack problem, which we solve by Lagrangian relaxation (\cref{alg:alloc}). Pricing each bit at $\lambda\ge0$ splits the objective into independent per-layer terms, so for a fixed $\lambda$ every layer simply takes
\begin{equation}
b_\ell(\lambda)=\arg\min_{b\in\mathcal{T}_\ell}\; S_\ell N_b+\lambda\,n_\ell B_b ,
\label{eq:lagrange}
\end{equation}
where $\mathcal{T}_\ell$ is the set of tiers allowed for layer $\ell$. A higher price moves layers to cheaper tiers, so the resulting average bits per weight $\mathrm{bpw}(\lambda)$ does not increase as $\lambda$ grows, and bisection finds the smallest price that meets the budget. Bits left over because tiers are discrete buy the upgrades with the largest loss reduction per bit.

\begin{algorithm}[t]
\caption{Bit allocation by Lagrangian relaxation (Stage~1)}
\label{alg:alloc}
\small
\begin{algorithmic}[1]
\Require sensitivities $S_\ell$, layer sizes $n_\ell$, allowed tiers $\mathcal{T}_\ell$, tier costs $B_b$ and noise $N_b$, budget $\bar B$ no smaller than the cost of the cheapest allowed tiers
\State $\mathrm{bpw}(\lambda)\gets\sum_\ell n_\ell B_{b_\ell(\lambda)}\big/\sum_\ell n_\ell$ \Comment{$b_\ell(\lambda)$ from \cref{eq:lagrange}}
\State $\lambda_{\mathrm{lo}}\gets0$;\ \ $\lambda_{\mathrm{hi}}\gets1$
\State \textbf{while} $\mathrm{bpw}(\lambda_{\mathrm{hi}})>\bar B$ \textbf{do} $\lambda_{\mathrm{hi}}\gets2\lambda_{\mathrm{hi}}$
\Repeat \Comment{bisection on the price}
  \State $\lambda\gets(\lambda_{\mathrm{lo}}+\lambda_{\mathrm{hi}})/2$
  \State \textbf{if} $\mathrm{bpw}(\lambda)>\bar B$ \textbf{then} $\lambda_{\mathrm{lo}}\gets\lambda$ \textbf{else} $\lambda_{\mathrm{hi}}\gets\lambda$
\Until{$\lambda_{\mathrm{hi}}-\lambda_{\mathrm{lo}}$ is negligible}
\State $b_\ell\gets b_\ell(\lambda_{\mathrm{hi}})$ for every layer $\ell$
\While{some layer's next allowed tier $b_\ell^{+}$ fits in the leftover bits $\bar B\sum_\ell n_\ell-\sum_\ell n_\ell B_{b_\ell}$} \Comment{spend the slack}
  \State upgrade the layer maximizing $S_\ell\,(N_{b_\ell}-N_{b_\ell^{+}})\,/\,\bigl(n_\ell\,(B_{b_\ell^{+}}-B_{b_\ell})\bigr)$
\EndWhile
\State \Return $\{b_\ell\}$
\end{algorithmic}
\end{algorithm}

Sensitive layers (large $S_\ell$ per weight) therefore receive more bits. One geometric consequence is worth noting: in the $(B_b,N_b)$ plane the 2-bit point lies above the straight line between the ternary and 3-bit points. Scaling the points by a layer's sensitivity and size does not change this, so for every price $\lambda$, \cref{eq:lagrange} prefers ternary or 3 bits to 2 bits for every layer and never chooses 2 bits on its own. In our runs only query and key projections are held at $\ge$2 bits; for Qwen3-1.7B at $\bar B=2.0$ the result is 75.3\% ternary, 6.0\% 2-bit and 18.7\% 3-bit weights. The greedy step of \cref{alg:alloc} can still move a ternary layer up to 2 bits when leftover bits allow; without the query--key rule this happens to three layers of Qwen3-1.7B. \Cref{sec:alloc} shows that holding query and key at $\ge$2 bits is unnecessary on this model.

\subsection{\texttt{Search}: How to best inititalize codewords with PTQ?}
\label{sec:stage2}
This stage of \texttt{ThinkBit} aims at initializing the codewords appropriately at the allocated precision using PTQ methods. Rounding each weight to its nearest codeword ignores how errors in different weights interact. GSQ~\citep{dadgarnia2026gsq} instead treats every codeword as a learnable distribution over candidate grid values $v_1,\dots,v_K$ with logits $z\in\mathbb{R}^K$, optimizes all distributions and scales of one transformer block jointly, and finally keeps the most likely value of each. It needs a starting point, a differentiable relaxations of the discrete choice presented in \Cref{tab:relax}, and an objective.

\paragraph{Layer-wise start.} Every 2- and 3-bit layer is first quantized on its own with GPTQ~\citep{frantar2023gptq}, which minimizes the layer's output error on calibration inputs,
\begin{equation}
\min_{\widehat W}\ \operatorname{tr}\bigl((W-\widehat W)\,H\,(W-\widehat W)^{\top}\bigr),\qquad H=H_X+\mu I,
\label{eq:gptq}
\end{equation}
by rounding one input column at a time and moving the not-yet-rounded columns to compensate. With $R$ the upper Cholesky factor of $H^{-1}$, after column $j$ is rounded its scaled error $e_j=(W_{:,j}-\widehat W_{:,j})/R_{jj}$ is subtracted as $W_{:,j'}\leftarrow W_{:,j'}-e_j\,R_{jj'}$ from every later column $j'>j$. The scale of each group is set when the group's first column is reached, from the current (already compensated) weights of each row $i$: $s=2\rho\max_j|W_{ij}|/(|\mathcal{C}_b|-1)$, with the maximum over the group's columns $j$ and the shrink factor $\rho\le1$ that minimizes the group's rounding error. For ternary this is $s=\rho\max_j|W_{ij}|$. Following KronQ's use of the output side, we also let the layer outputs set the damping $\mu$: with $\phi_o=\mathbb{E}_t[y_{t,o}^2]$ the second moment of output channel $o$ and $\bar\phi$ its mean over channels,
\begin{equation}
\mu=\mu_0\,\overline{\operatorname{diag}(H_X)}\cdot\operatorname{clip}\Bigl(\frac{1}{d_{\mathrm{out}}}\sum_{o}\sqrt{\phi_o/\bar\phi}\Bigr),
\label{eq:damping}
\end{equation}
where $\mu_0$ is a base factor, $\overline{\operatorname{diag}(H_X)}$ is the mean diagonal entry and the clip bounds the factor to a fixed range. Because the ratios $\phi_o/\bar\phi$ average to one, the factor is at most one (Jensen's inequality) and falls as the output energy concentrates in fewer channels, so such layers receive less damping; in practice only the lower bound of the clip can bind. All other 2- and 3-bit layers keep the GPTQ codewords as their starting codewords $c^0$. Query and key projections, at any tier, refine their GPTQ codewords for a fixed number of steps with the relaxations below, applied to \cref{eq:gptq}, keep the result, and skip the block search. Ternary layers skip GPTQ: they run the same layer-wise refinement from uninformative logits, with GPTQ's scale rule, and the result is their starting point for the block search.

\paragraph{Gumbel-softmax relaxation (3-bit and ternary layers).} For a 3-bit weight, the candidates are the starting codeword and its neighbors, $v_k=c^0+\delta_k$ with $\delta_k\in\{-2,\dots,2\}$ (restricted to the grid), and the logits start proportional to $-\tfrac12\delta_k^2$ (plus small noise) so that $c^0$ is the most likely value. Each forward pass draws Gumbel noise $\epsilon_k$ and uses the $\pi$-weighted average of the candidates, where $\pi$ is the resulting soft one-hot vector~\citep{jang2017gumbel}:
\begin{equation}
\begin{gathered}
\pi_k=\frac{\exp\bigl((\kappa z_k+\epsilon_k)/\tau\bigr)}{\sum_{k'}\exp\bigl((\kappa z_{k'}+\epsilon_{k'})/\tau\bigr)},\qquad \epsilon_k=-\log(-\log U_k),\quad U_k\sim\mathcal{U}(0,1),\\
\tilde c=\sum_k\pi_k v_k,\qquad \frac{\partial\tilde c}{\partial z_k}=\frac{\kappa}{\tau}\,\pi_k\,(v_k-\tilde c).
\end{gathered}
\label{eq:gumbel}
\end{equation}
The Gumbel noise makes $\pi$ a random sample whose largest entry is distributed as $\operatorname{softmax}(\kappa z)$, and the gradient is taken for the drawn noise; the temperature $\tau$ controls how close $\pi$ is to one-hot and the sharpening factor $\kappa$ scales the logits. Both are annealed over training, so the soft codeword $\tilde c$ starts as a blend of neighbors and ends near a single grid value. A ternary codeword factors into a mask and a sign, $c=m\,\sigma$ with $m\in\{0,1\}$ and $\sigma\in\{-1,1\}$, and each factor is a binary relaxation (the binary concrete distribution~\citep{maddison2017concrete}) with its own logit, $z^{m}$ or $z^{\sigma}$:
\begin{equation}
\tilde m=\operatorname{sigmoid}\Bigl(\frac{2\kappa z^{m}+\eta}{\tau}\Bigr),\qquad
\tilde\sigma=2\operatorname{sigmoid}\Bigl(\frac{2\kappa z^{\sigma}+\eta'}{\tau}\Bigr)-1,\qquad
\tilde c=\tilde m\,\tilde\sigma,\qquad \eta,\eta'\sim\operatorname{Logistic}(0,1).
\label{eq:ternary}
\end{equation}
At the end the codeword is rounded to $m=\mathbf{1}[z^{m}>0]$ and $\sigma=\operatorname{sign}(z^{\sigma})$.

\paragraph{ReinMax (2-bit layers).} On the 2-bit grid $\{-2,-1,0,1\}$, each forward pass instead draws a \emph{hard} sample: with probabilities $q=\operatorname{softmax}(\kappa z)$, a one-hot vector $\omega\sim\operatorname{Cat}(q)$ selects the codeword $c=\sum_k \omega_k v_k$, where the candidates $v_k$ are now the four grid values, so the forward pass uses exactly the kind of weight that inference will use. A discrete sample has no gradient, so ReinMax~\citep{liu2023reinmax} estimates $\nabla_z\,\mathbb{E}_\omega[\mathcal{L}]$. Writing $\nabla_\omega=\partial\mathcal{L}/\partial\omega$ and $J(\pi)=\operatorname{diag}(\pi)-\pi\pi^{\top}$ for the Jacobian of a softmax with output $\pi$,
\begin{equation}
\pi_{1/2}=\tfrac12\bigl(\operatorname{softmax}(\kappa z/\tau)+\omega\bigr),\qquad
\widehat\nabla_{\kappa z}\mathcal{L}=2\,J(\pi_{1/2})\,\nabla_\omega-\tfrac12\,J(q)\,\nabla_\omega ,
\label{eq:reinmax}
\end{equation}
centered to zero mean over $k$. The straight-through Gumbel estimator would use only $\tau^{-1}J(\pi)\,\nabla_\omega$ with $\pi$ from \cref{eq:gumbel}, a first-order approximation; ReinMax evaluates the Jacobian at the midpoint $\pi_{1/2}$ between the soft probabilities and the drawn sample, as in Heun's method for ordinary differential equations, which makes the estimate accurate to second order. It requires $\tau\ge1$, so the annealed temperature is clamped at 1.

\begin{table}[t]
\centering
\small
\setlength{\tabcolsep}{4pt}
\begin{tabular}{L{2.6cm}L{5.2cm}L{5.2cm}}
\toprule
 & Gumbel-softmax (\cref{eq:gumbel,eq:ternary}) & ReinMax (\cref{eq:reinmax}) \\
\midrule
Forward weight & soft blend of grid values, $\tilde c=\sum_k\pi_kv_k$ & hard grid value, $c=\sum_k\omega_kv_k$ \\
Gradient & exact gradient of the soft weight & second-order estimate for the hard weight \\
Temperature $\tau$ & annealed toward 0 & kept $\ge1$ \\
Train--test gap & vanishes only as $\pi$ becomes one-hot & none: training uses grid weights \\
Used for & ternary and 3-bit layers & 2-bit layers \\
\bottomrule
\end{tabular}
\caption{The two relaxations used by codeword search. Gumbel-softmax trains a smooth stand-in for the rounded weight with an exact gradient; ReinMax trains the rounded weight itself with an approximate gradient.}
\label{tab:relax}
\end{table}

\paragraph{Block objective.} The relaxations are trained on the output of each transformer block rather than each layer, so codewords can compensate for errors elsewhere in the block. Within a block, GSQ first fits the attention projections against the residual stream after attention and then the multi-layer perceptron (MLP) projections against the block output. Inputs are the activations of the already quantized preceding blocks, so later blocks also correct the errors of earlier ones, and the target is the \texttt{bf16} block applied to the same inputs. With $\hat a_t$ and $a_t$ the quantized and \texttt{bf16} residual-stream vectors of width $d_{\mathrm{model}}$ at token $t$, over $T$ calibration tokens,
\begin{equation}
\mathcal{L}_{\mathrm{blk}}=\frac{1}{T\,d_{\mathrm{model}}}\sum_{t=1}^{T}\sum_{o=1}^{d_{\mathrm{model}}}w_o\,(\hat a_{t,o}-a_{t,o})^2,\qquad w=\operatorname{norm}\bigl(\operatorname{diag}H_G^{\mathrm{res}}\bigr),
\label{eq:block}
\end{equation}
where $H_G^{\mathrm{res}}$ is the gradient second moment of \cref{eq:kronq} measured at that point of the residual stream, so channels whose errors move the loss more are weighted more. $\operatorname{norm}(\cdot)$ rescales the channel weights $w_o$ to mean 1 and bounds them so that no channel is ignored or dominates. Because the change of a layer's output at token $t$ is $E_\ell x_t$, \cref{eq:dloss} equals $\tfrac12\mathbb{E}_t[(E_\ell x_t)^{\top}H_{G,\ell}(E_\ell x_t)]$, a gradient-weighted output error; \cref{eq:block} is a normalized, bounded version of this quantity with a diagonal $H_G$, measured on the block's output rather than on a single layer's. Logits and scales are trained jointly for a fixed number of epochs per block, after which every codeword takes its most likely value.

\subsection{\texttt{Recalibrate}: Is tuning scales and norms enough?}
\label{sec:stage3}
Fine-tuning only the higher-precision parameters, the scales and the RMSNorm gains, is cheap since they make up less than 1\% of the model (one scale per 128 weights) and hence the optimizer state stays small. This stage of \texttt{ThinkBit} applies recalibration to scales and norm gains to measure how much accuracy it recovers on its own. Search fits each block in isolation, so small errors still add up across the model. Scale recalibration repairs them without touching the codewords, and thus without changing the stored integers. Every group scale is reparameterized as $s=s_0e^{r}$ and every RMSNorm gain as $\nu=\nu_0(1+\psi)$, where $s_0$ and $\nu_0$ are the values after Stage~2 and the offsets $r$ and $\psi$ are bounded. Only the offsets $r$ and $\psi$ are trained, so the model keeps exactly the same format and size. They minimize the forward Kullback--Leibler (KL) divergence from the \texttt{bf16} model's next-token distribution $p_{\texttt{bf16}}$ to the quantized model's $p_{r,\psi}$~\citep{hinton2015distilling}, with small penalties that keep scales and gains near their starting values:
\begin{equation}
\mathcal{L}_{\mathrm{rec}}=\frac{1}{T}\sum_{t=1}^{T}\operatorname{KL}\bigl(p_{\texttt{bf16}}(\cdot\mid \xi_{\le t})\,\big\|\,p_{r,\psi}(\cdot\mid \xi_{\le t})\bigr)+w_r\,\overline{r^2}+w_\psi\,\overline{\psi^2},
\label{eq:recal}
\end{equation}
where $\overline{\,\cdot\,}$ averages over all scales or gains and $w_r,w_\psi$ weight the penalties. The penalties act on the offsets, $r=\log(s/s_0)$ and $\psi=\nu/\nu_0-1$, not on the magnitudes of $s$ and $\nu$: a scale or gain that keeps its Stage-2 value costs nothing, and a given relative change costs the same at any magnitude. They act as a trust region around the Stage-2 solution, whose codewords were chosen jointly with $s_0$ and stay frozen here. Unlike the block loss, the KL is measured on the whole model's output, so a scale can trade a slightly larger local error for a smaller error downstream. Training keeps the step with the lowest held-out KL.

\subsection{\texttt{QAT}: Does moving codewords help?}
\label{sec:stage4}
QAT lets codewords move, giving the model more freedom than recalibration. This stage of \texttt{ThinkBit} measures what that freedom adds, and whether it pays for its cost. Each weight gets a 32-bit floating-point (FP32) latent $u_{ij}$ measured in codeword units. The forward pass rounds it onto the layer's grid $[c_{\min},c_{\max}]$, and the backward pass treats the rounding as the identity (the straight-through estimator, STE~\citep{bengio2013ste}):
\begin{equation}
\widehat W_{ij}=s_{i,\gamma(j)}\bigl(\bar u_{ij}+\operatorname{sg}\bigl[\operatorname{round}(\bar u_{ij})-\bar u_{ij}\bigr]\bigr),\qquad \bar u_{ij}=\operatorname{clip}\bigl(u_{ij},\,c_{\min}-\beta,\,c_{\max}+\beta\bigr),
\label{eq:ste}
\end{equation}
where $\operatorname{sg}$ stops the gradient and $\beta$ is just below $\tfrac12$, so $\partial\widehat W_{ij}/\partial u_{ij}=s_{i,\gamma(j)}$ inside the clip range. A codeword is reassigned when its latent crosses a rounding boundary, halfway between two grid values. If every latent started exactly on its codeword, small gradient steps would rarely cross a boundary. Latents therefore start partway toward the \texttt{bf16} weight,
\begin{equation}
u_{ij}^{(0)} = c_{ij} + \operatorname{clip}\!\bigl(\alpha\,(W_{ij}/s_{i,\gamma(j)} - c_{ij}),\ -\beta,\ \beta\bigr),
\label{eq:residual}
\end{equation}
where $W_{ij}$ is the \texttt{bf16} weight and $\alpha\in[0,1]$ sets how far toward it the latent starts ($\alpha=0.5$ in the framework). This leaves every codeword unchanged at step 0 (the offset stays below half a step) but places each latent on the side of its codeword where the \texttt{bf16} weight lies, so training can move it across the boundary when the loss favors it. Scales and gains keep training. The loss adds a hidden-state distillation term to the forward KL of \cref{eq:recal}:
\begin{equation}
\mathcal{L}_{\mathrm{QAT}}=\frac{1}{T}\sum_{t=1}^{T}\Bigl[\operatorname{KL}\bigl(p_{\texttt{bf16}}\,\big\|\,p_{u}\bigr)_t+w_h\,\frac{1}{|\mathcal{S}|}\sum_{k\in\mathcal{S}}\frac{\|\hat h_{k,t}-h_{k,t}\|_2^2}{\|h_{k,t}\|_2^2}\Bigr],
\label{eq:qatloss}
\end{equation}
where $p_u$ is the quantized model's next-token distribution, $h_{k,t}$ and $\hat h_{k,t}$ are the \texttt{bf16} and quantized outputs of transformer block $k$ at token $t$, $\mathcal{S}$ is a subset of blocks that includes the last, and $w_h$ weights the term. QAT can also quantize the output head and the embedding to 2 bits with the same latents; for models whose head and embedding share weights (such as Qwen3-1.7B), it first unties them. As in Stage~3, training keeps the step with the lowest held-out KL.

\subsection{\texttt{Selection}: Which checkpoint reasons better, recalibrated or QAT?}
\label{sec:stage5}
Stages~3 and~4 produce two candidates, the recalibrated checkpoint (\ck{Rec}) and the QAT checkpoint (\ck{QAT}), and likelihood metrics do not tell which reasons better (\cref{sec:controls}). The stage of \texttt{ThinkBit}, therefore compares them on $P$ held-out reasoning problems: level-3--5 MATH training problems, disjoint from MATH-500 (for example, $P=313$). Each candidate samples four answers per problem in thinking mode, and an answer counts only if the reasoning trace closes and the final answer is correct (strict scoring). Let $\mathrm{acc}^A_i$ and $\mathrm{acc}^B_i$ be the fractions of correct answers of candidates $A$ and $B$ on problem $i$, and $\Delta_i=\mathrm{acc}^A_i-\mathrm{acc}^B_i$. The paired sign-flip test gives
\begin{equation}
p=\Pr_{\chi}\Bigl[\,\Bigl|\sum_{i=1}^{P}\chi_i \Delta_i\Bigr|\ \ge\ \Bigl|\sum_{i=1}^{P}\Delta_i\Bigr|\,\Bigr],\qquad \chi_i\in\{-1,+1\}\ \text{independent and uniform},
\label{eq:signflip}
\end{equation}
estimated from 20,000 random sign vectors (counting the observed signs once, so $p>0$). The rule keeps the candidate with the higher mean accuracy if $p\le0.05$ and otherwise the smaller checkpoint, or the recalibrated one when both have the same size.

\section{Experimental setup}
\label{sec:data}

\paragraph{Models.} We primarily study \texttt{ThinkBit} on the Qwen-3 family. Specifically, Qwen3-1.7B, Qwen3-8B and Qwen3-14B~\citep{yang2025qwen3} (thinking mode, native tool calling; Qwen3-8B and Qwen3-14B have untied heads). For all models we use the same hyperparameters for \texttt{ThinkBit}.
\paragraph{Data.} The quantization and training data mixture has three parts (i)~each model's \emph{own} \texttt{bf16} replies to prompts, namely Magicoder code instructions, GSM8K-train and MetaMathQA math problems, Hermes and Glaive tool calls, TriviaQA, Natural Questions and WebQuestions training questions, and MMLU-auxiliary questions without answers; (ii)~the model's own continuations of FineWeb-Edu and Wikipedia passages; and (iii)~raw text from FineWeb-Edu, FineWeb, Wikipedia (WikiText articles removed), OpenWebMath, Evol-Instruct code, Hermes and Glaive multi-turn tool conversations and Python source (codeparrot-clean). Replies are sampled with each model's chat template, which for Qwen3 includes the thinking block, and are capped at 4,096 tokens, so training never sees a complete long reasoning trace. \Cref{tab:data} lists the data and amount used by each stage.

\paragraph{Reasoning data for the final QAT.} The stage studies (\cref{sec:stages,sec:controls}) use the mixture above. The selected checkpoints of \cref{sec:scaling,sec:storage} repeat Stage~4 for longer on a mixture that adds complete long reasoning traces: each model's own \texttt{bf16} thinking-mode answers to public math and code prompts and to hard competition problems (OpenMathReasoning and OpenCodeReasoning statements; truncated traces dropped), DeepSeek-R1 solutions from OpenMathReasoning, and PG-19 books and arXiv papers for long context; at 1.7B also OpenR1-Math-220k, Codeforces CoTs, OpenCodeReasoning, OpenCodeInstruct and Cosmopedia textbooks. Reasoning traces make up about 60\% of the mixture. Every source passes the same decontamination against all evaluated benchmarks as the base mixture. At 1.7B, \ck{QAT+R-h2} starts from \ck{Rec-h2} (2.1B tokens, learning rate $2\cdot10^{-3}$); at 8B and 14B it starts from a further recalibration of \ck{Rec-h2} on the long-trace data (1.1B tokens, $1.5\cdot10^{-3}$); format and bytes are unchanged.

\paragraph{Checkpoint names.} We name every checkpoint by the stages it went through (\cref{tab:names}): a base name for the codeword start and the trained stages, a suffix for the precision of the head and embedding, and \ck{+R} when Stage~4 (or~3) used the reasoning mixture. ``Selected'' marks the checkpoint kept by Stage~5. The same names are used in every table, caption and figure.

\begin{table}[t]
\centering
\small
\setlength{\tabcolsep}{4pt}
\begin{tabular}{lL{10.2cm}}
\toprule
Name & Checkpoint \\
\midrule
\ck{GSQ} & Stages~1--2: tiers allocated and codewords searched; no training \\
\ck{Rec} & \ck{GSQ} + Stage~3: scales and RMSNorm gains recalibrated, codewords frozen \\
\ck{QAT} & \ck{Rec} + Stage~4: QAT reassigns codewords (the full framework) \\
\ck{QAT-noRec} & \ck{GSQ} + Stage~4, without recalibration \\
\ck{LW-Rec}, \ck{LW-QAT} & as \ck{Rec} and \ck{QAT}, but starting from the layer-wise PTQ solution instead of codeword search \\
\ck{QAT-only} & layer-wise start + Stage~4 only: no codeword search, no recalibration \\
\midrule
\ck{-hbf16} & suffix: head and embedding kept in \texttt{bf16} (tied at 1.7B) \\
\ck{-h2} & suffix: head and embedding quantized to 2 bits (untied at 1.7B) \\
\ck{+R} & trained on the reasoning mixture of \cref{sec:data}, e.g.\ \ck{Rec+R-h2}, \ck{QAT+R-h2} \\
\bottomrule
\end{tabular}
\caption{Checkpoint names. At 1.7B, \ck{Rec-h2} uses a second recalibration pass that quantizes the head and embedding to 2-bits; \ck{Rec-h2} (1 pass) uses one.}
\label{tab:names}
\end{table}

\begin{table}[t]
\centering
\small
\setlength{\tabcolsep}{4pt}
\begin{tabular}{L{3.3cm}L{6.4cm}r}
\toprule
Stage & Data & Amount \\
\midrule
1: KronQ sensitivity & subset of the Stage-2 calibration data & 2M tokens \\
2: Codeword search (GSQ) & own replies, tool conversations, code and FineWeb-Edu & 57M tokens \\
3: Scale recalibration & all sources (i)--(iii) presented in Section \ref{sec:data} & 268M tokens \\
4: QAT & all sources (i)--(iii) presented in Section \ref{sec:data} & 537M tokens \\
4$^\ast$: QAT, reasoning data (\ck{+R}) & (i)--(iii) plus complete long reasoning traces (\cref{sec:data}) & 1.1--2.1B tokens \\
5: Checkpoint selection & level-3--5 MATH training problems, disjoint from MATH-500 & 313 problems \\
\bottomrule
\end{tabular}
\caption{Data used by each stage. $^\ast$Longer QAT on the reasoning mixture, used for the selected checkpoints.}
\label{tab:data}
\end{table}

\paragraph{Evaluation.} Dense \texttt{bf16} exports are scored with the Language Model Evaluation Harness (lm-eval)~\citep{eval-harness} and served by vLLM. The core suite contains ARC-e/c, MMLU, WinoGrande, HellaSwag, PIQA and LAMBADA; their mean log-likelihood accuracy is the \emph{7-task mean}. Generation covers math, code, instruction following and tool use on the Berkeley Function Calling Leaderboard (BFCL)~\citep{patil2025bfcl}. The broad suite adds MMLU-Pro, TriviaQA, TruthfulQA, SciQ, OpenBookQA, CommonsenseQA, BoolQ, RACE, Beyond the Imitation Game Benchmark Hard (BBH), MATH-500, GSM8K chain-of-thought (GSM8K-CoT), Instruction Following Evaluation (IFEval), Mostly Basic Python Problems (MBPP) and HumanEval+. American Invitational Mathematics Examination (AIME) 2024/2025 and MATH-500 use the models' chat/thinking format; We report thinking-mode MATH-500 in two ways. \emph{Greedy} decodes each problem once and grades it with lm-eval. \emph{Sampled} draws four answers per problem (temperature 0.6, top-$p$ 0.95, top-$k$ 20, up to 32k new tokens) and scores them \emph{strictly}: an answer counts only if the reasoning trace closes with \texttt{</think>} and the final answer is correct. Sampled, strictly scored MATH-500 is our primary reasoning metric. AIME@16 denotes mean sampled pass@1 over 16 samples per problem.

\section{Where does the quality come from?}
\label{sec:stages}
\begin{figure}[t]
\centering
\includegraphics[width=\linewidth]{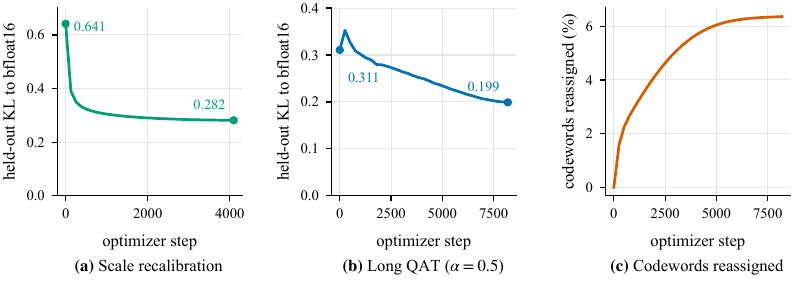}
\caption{Qwen3-1.7B training curves. \textbf{(a)}~Scale recalibration (codewords frozen): held-out KL to \texttt{bf16} falls from 0.641 to 0.282. \textbf{(b)}~Long QAT from the recalibrated model: KL increases briefly while codewords start to move, then falls from 0.311 to 0.199 (QAT is scored on the text that follows its own training slice, so its starting KL differs from the end of (a)). \textbf{(c)}~Share of integer codewords that differ from the recalibrated model.}
\label{fig:recqat}
\end{figure}

We follow Qwen3-1.7B through Stages 3 and 4 (\cref{tab:stages17,fig:recqat}) and then split reasoning accuracy into finishing a trace and solving the problem (\cref{tab:decomp}).

\begin{table}[t]
\centering
\small
\setlength{\tabcolsep}{4pt}
\begin{tabular}{lcccccc}
\toprule
Qwen3-1.7B & WikiText-2 $\downarrow$ & 7-task & MMLU & GSM8K & HumanEval & MATH-500 \\
\midrule
\texttt{bf16} & 16.67 & 58.9 & 55.5 & 68.6 & 41.5 & 91.2 \\
Stage 2: \ck{GSQ-hbf16} & 46.91 & 43.8 & 36.4 & 27.5 & 8.5 & 14.8 \\
\rowcolor{ourrow}Stage 3: \ck{Rec-hbf16} & 19.46 & 51.5 & 44.4 & \textbf{51.6} & \textbf{22.6} & \textbf{47.5} \\
\rowcolor{ourrow}Stage 4: \ck{QAT-hbf16} & \textbf{16.54} & \textbf{54.6} & \textbf{49.9} & 50.2 & 17.7 & 43.0 \\
\bottomrule
\end{tabular}
\caption{Recalibration recovers most of what codeword search loses; QAT further improves likelihood but not reasoning. Qwen3-1.7B after each stage (\texttt{bf16} head and embedding); Bold: best quantized value.}
\label{tab:stages17}
\end{table}

\paragraph{Recalibration closes most of the likelihood gap.}
With the Stage 2 codewords frozen, recalibration lowers WikiText-2 perplexity from 46.91 to 19.46 (\texttt{bf16}: 16.67), raises the 7-task mean from 43.8 to 51.5 (\texttt{bf16}: 58.9) and sampled MATH-500 from 14.8 to 47.5 (\Cref{tab:stages17}). The scales do almost all of the work: held-out KL is 0.641 after GSQ, 0.282 with the scales trained, 0.418 with only the RMSNorm gains trained and 0.282 with both (\Cref{tab:recmech} in \Cref{app:recmech}).

\paragraph{The scales fit the output, not the weights.}
Recalibration moves each scale by a median of 6\% (log-offset $|r|=0.062$ in $s=s_0e^{r}$, \Cref{sec:stage3}), but not toward the scale that best fits the frozen codewords in weight space, $s^\star_g=\langle W_g,c_g\rangle/\langle c_g,c_g\rangle$ for group $g$: $r_g$ has the same sign as $\log(s^\star_g/s_{0,g})$ in 51\% of groups, and the two correlate at 0.07. Relative weight error $\|W-s\,c\|^2/\|W\|^2$ therefore improves slightly, from 0.386 to 0.393, while $s^\star$ would lower it to 0.277. The objective explains why. \Cref{eq:recal} scores the whole network's output, so it weights each weight error by the activations it meets and lets the scales of one layer offset rounding errors in others. Block outputs show the same pattern (\Cref{tab:scalestats} in \Cref{app:recmech}). Given the same \texttt{bf16} input, a recalibrated block is \emph{further} from its \texttt{bf16} output than before (mean relative error 0.094 vs.\ 0.079), and the hidden states of the recalibrated network are further from \texttt{bf16} on average (0.28 vs.\ 0.21), yet next-token KL on WikiText-2 falls from 1.30 to 0.57. Recalibration does not make intermediate representations more faithful; it adjusts them so that the output distribution matches.

\paragraph{QAT reassigns few codewords.}
Long running QAT reassigns 6.4\% of the codewords (\Cref{fig:recqat}c), each by a single grid step toward its \texttt{bf16} weight (\Cref{app:moves}); initialization, not learning rate, decides how many move (\cref{app:factorial}). It brings WikiText-2 to 16.54, partly through Wikipedia in its training mix, the 7-task mean to 54.6 and MMLU from 44.4 to 49.9. GSM8K, HumanEval and sampled MATH-500 do not improve; they fall from 51.6, 22.6 and 47.5 to 50.2, 17.7 and 43.0 (\Cref{tab:stages17}). On a broader suite, reading comprehension and commonsense keep 80\% or more of \texttt{bf16}'s score, while code and non-chat math do not recover (\Cref{app:thorough}).
\begin{table}[t]
\centering
\small
\begin{tabular}{llccc}
\toprule
Model & Checkpoint & Strict acc. & Closed (\%) & Acc.\ on closed \\
\midrule
Qwen3-1.7B & \texttt{bf16} & 91.2 & 99.2 & 91.9 \\
\rowcolor{ourrow} & \ck{Rec-hbf16} & \textbf{47.5} & \textbf{88.0} & 53.9 \\
\rowcolor{ourrow} & \ck{QAT-hbf16} & 43.0 & 69.6 & 61.7 \\
 & \ck{QAT-noRec-hbf16} & 37.0 & 57.2 & \textbf{64.5} \\
 & \ck{LW-QAT-hbf16} & 35.0 & 68.2 & 51.3 \\
 & \ck{QAT-only-hbf16} & 32.1 & 55.4 & 58.0 \\
\midrule
Qwen3-14B & \texttt{bf16} & 97.5 & 99.9 & 97.6 \\
\rowcolor{ourrow} & \ck{Rec-hbf16} & \textbf{92.8} & 97.3 & \textbf{95.4} \\
 & \ck{LW-Rec-hbf16} & 89.7 & \textbf{97.5} & 91.9 \\
\rowcolor{ourrow} & \ck{QAT-h2} & 90.5 & 96.8 & 93.6 \\
 & \ck{QAT-only-h2} & 89.6 & 96.4 & 92.9 \\
\bottomrule
\end{tabular}
\caption{Two-bit models fail mainly by not finishing their reasoning; at 14B almost every trace finishes. Sampled MATH-500 split into the share of traces that close \texttt{</think>} and the accuracy on those. Bold: best quantized value per model.}
\label{tab:decomp}
\end{table}

\paragraph{Finishing reasoning traces versus solving the task.}
Strict scoring counts an answer only if it follows a closed \texttt{</think>} tag, so a two-bit model can fail by looping as well as by reasoning wrongly. \Cref{tab:decomp} separates the two: \texttt{bf16} Qwen3-1.7B closes 99.2\% of its traces, the two-bit checkpoints of \Cref{tab:decomp} 55.4--88.0\% and the 2-bit GGUFs 47--56\%; budget forcing narrows but does not close this gap (\Cref{app:budget}). Codeword search raises accuracy on closed traces, from 51.3 to 61.7 on Qwen3-1.7B, while the closing rate changes less (68.2 to 69.6; problem-matched tests in \Cref{tab:matched}). QAT and recalibration act mainly on closing: QAT lowers it from 88.0 to 69.6 on Qwen3-1.7B, in the direction of its total effect, and recalibration raises it from 57.2 to 69.6 on Qwen3-1.7B. At 14B almost every trace closes (96.4--97.5\%), so codeword search shows up only in closed-trace accuracy (95.4 vs.\ 91.9).

\section{Scaling across model sizes}
\label{sec:scaling}
\texttt{ThinkBit} with every hyper-parameter fixed on Qwen3-1.7B, was applied unchanged to Qwen3-8B and Qwen3-14B. \Cref{tab:scaling} compares each selected checkpoint, \ck{QAT+R-h2} (QAT on the reasoning mixture of \cref{sec:data}), with its \texttt{bf16} model and with the smallest and largest published 2-bit GGUF files of the same model.

\providecommand{\gr}[1]{\textcolor{gray}{#1}}
\begin{table}[!htbp]
\centering
\small
\setlength{\tabcolsep}{4pt}
\resizebox{\linewidth}{!}{%
\begin{tabular}{llcccccccc}
\toprule
Model & Checkpoint & GB & bits/param & 7-task & MMLU & GSM8K & HumanEval & MATH-500 & AIME@16 \\
\midrule
Qwen3-1.7B & \texttt{bf16} & 3.44 & 16 & 58.9 & 55.5 & 68.6 & 41.5 & 91.2 & 41.4 \\
 & \multicolumn{9}{l}{\emph{$\le$2.6 bits per parameter}} \\
\rowcolor{ourrow} & \ck{QAT+R-h2} (selected)$^\S$ & 0.52 & 2.42 & 53.8 & 47.0 & 58.7 & 25.6 & \textbf{73.0} & \textbf{12.3} \\
 & \ck{Rec+R-h2}$^\S$ & 0.52 & 2.42 & 50.6 & 43.5 & 51.7 & 22.0 & 55.6 & 3.0 \\
 & \ck{Rec-h2} & 0.52 & 2.42 & 51.1 & 45.2 & 51.8 & 22.6 & 46.7 & 1.0 \\
 & 1-bit Bonsai & 0.25 & 1.15 & 49.7 & 41.7 & 43.9 & 37.2 & 28.7 & 0.7 \\
 & Ternary Bonsai & 0.46 & 2.15 & \textbf{55.0} & \textbf{48.6} & \textbf{60.0} & \textbf{48.2} & 46.4 & 1.4 \\
 & EfficientQAT, 2-bit head/emb.$^\ddagger$ & 0.54 & 2.53 & 51.9 & 45.8 & 43.7 & 22.0 & 30.0 & 0.6 \\
 & BitDistiller, 2-bit head/emb.$^\ddagger$ & 0.54 & 2.53 & 51.0 & 45.6 & 42.4 & 16.5 & 25.3 & 0.1 \\
 & \multicolumn{9}{l}{\emph{2.6--4 bits per parameter}} \\
 & GGUF UD-IQ2\_XXS~\citep{unsloth-qwen3-1.7b-udiq2xxs} & 0.61 & 2.82 & 38.2 & 26.0 & 1.3 & 1.2 & 1.0 & 0.0 \\
 & GGUF Q2\_K~\citep{unsloth-qwen3-1.7b-q2k} & 0.78 & 3.62 & 49.2 & 40.0 & 31.8 & 18.3 & 41.9 & 0.1 \\
 & \multicolumn{9}{l}{\emph{$>$4 bits per parameter}} \\
 & EfficientQAT$^\dagger$ & 1.00 & 4.65 & 53.0 & 46.2 & 42.3 & 22.6 & 34.2 & 1.0 \\
 & BitDistiller$^\dagger$ & 1.00 & 4.65 & 51.7 & 45.9 & 41.4 & 20.7 & 27.8 & 0.1 \\
\midrule
Qwen3-8B & \texttt{bf16} & 16.4 & 16 & 70.7 & 73.0 & 87.9 & 63.4 & 96.9 & 71.2 \\
 & \multicolumn{9}{l}{\emph{$\le$2.6 bits per parameter}} \\
\rowcolor{ourrow} & \ck{QAT+R-h2} (selected)$^\S$ & 2.08 & 2.03 & 65.8 & \textbf{64.8} & \textbf{84.5} & 47.0 & \textbf{92.2} & \textbf{46.9} \\
 & \ck{Rec+R-h2}$^\S$ & 2.08 & 2.03 & 65.1 & 63.8 & 83.2 & 56.1 & 89.3 & 38.4 \\
 & \ck{Rec-h2} & 2.08 & 2.03 & 64.9 & 63.7 & 82.8 & 53.0 & 86.6 & 28.8 \\
 & 1-bit Bonsai & 1.16 & 1.13 & 62.0 & 57.8 & 77.8 & 61.0 & 71.8 & 12.6 \\
 & Ternary Bonsai & 2.18 & 2.13 & \textbf{66.5} & 64.4 & 82.8 & \textbf{62.2} & 83.7 & 24.4 \\
 & GGUF UD-IQ2\_XXS~\citep{unsloth-qwen3-8b-udiq2xxs} & 2.61 & 2.54 & 60.4 & 59.2 & 47.4 & 24.4 & 35.2 & 1.0 \\
 & \multicolumn{9}{l}{\emph{2.6--4 bits per parameter}} \\
 & GGUF Q2\_K~\citep{unsloth-qwen3-8b-q2k} & 3.28 & 3.21 & 67.2 & 67.4 & 82.8 & 56.7 & 91.1 & 45.6 \\
 & \multicolumn{9}{l}{\emph{$>$4 bits per parameter}} \\
 & EfficientQAT$^\dagger$ & 4.35 & 4.25 & 66.5 & 62.4 & 72.5 & 33.5 & 71.9 & 10.3 \\
 & BitDistiller$^\dagger$ & 4.35 & 4.25 & 65.4 & 62.6 & 75.1 & 48.8 & 66.4 & 7.1 \\
\midrule
Qwen3-14B & \texttt{bf16} & 29.5 & 16 & 74.3 & 77.1 & 88.2 & 57.3 & 97.5 & 73.6 \\
 & \multicolumn{9}{l}{\emph{$\le$2.6 bits per parameter}} \\
\rowcolor{ourrow} & \ck{QAT+R-h2} (selected)$^\S$ & 3.74 & 2.03 & \textbf{69.9} & \textbf{70.5} & 87.3 & 59.8 & \textbf{94.8} & \textbf{57.7} \\
 & \ck{Rec+R-h2}$^\S$ & 3.74 & 2.03 & 69.5 & 69.7 & \textbf{88.2} & 59.1 & 94.4 & 53.5 \\
 & \ck{Rec-h2} & 3.74 & 2.03 & \textbf{69.9} & 69.5 & 87.7 & \textbf{65.2} & 92.7 & 46.5 \\
 & GGUF UD-IQ2\_XXS~\citep{unsloth-qwen3-14b-udiq2xxs} & 4.47 & 2.42 & 66.7 & 65.2 & 71.2 & 39.0 & 85.2 & 26.7 \\
 & \multicolumn{9}{l}{\emph{2.6--4 bits per parameter}} \\
 & GGUF Q2\_K~\citep{unsloth-qwen3-14b-q2k} & 5.75 & 3.12 & 71.6 & 73.8 & 87.3 & 54.9 & 95.0 & 63.0 \\
 & EfficientQAT$^\dagger$ & 6.65 & 3.60 & 71.2 & 68.7 & 77.5 & 45.1 & 85.4 & 28.4 \\
 & BitDistiller$^{\dagger\P}$ & 6.65 & 3.60 & 69.8 & 68.9 & 83.2 & 56.1 & 84.9 & 24.7 \\
\bottomrule
\end{tabular}}
\caption{At every size, \ck{QAT+R-h2} is the most accurate checkpoint on MATH-500 and AIME@16 among files of at most 2.6 bits per parameter. MATH-500: sampled; AIME@16: mean pass@1 over 16 samples. Bonsai \citep{prismml2026bonsai} and GGUF \citep{unsloth2025qwen3gguf} files run in our harness. Rows grouped by bits per parameter; bold: best in the $\le$2.6-bit group. $^\S$Trained on the reasoning mixture. $^\dagger$Official code on our base mixture, same tokens as \ck{Rec-h2}, \texttt{bf16} head and embedding. $^\ddagger$Same with a 2-bit head and embedding. $^\P$Trained with 4,096-token sequences.}
\label{tab:scaling}
\end{table}

\begin{figure}[!htbp]
\centering
\includegraphics[width=0.94\linewidth]{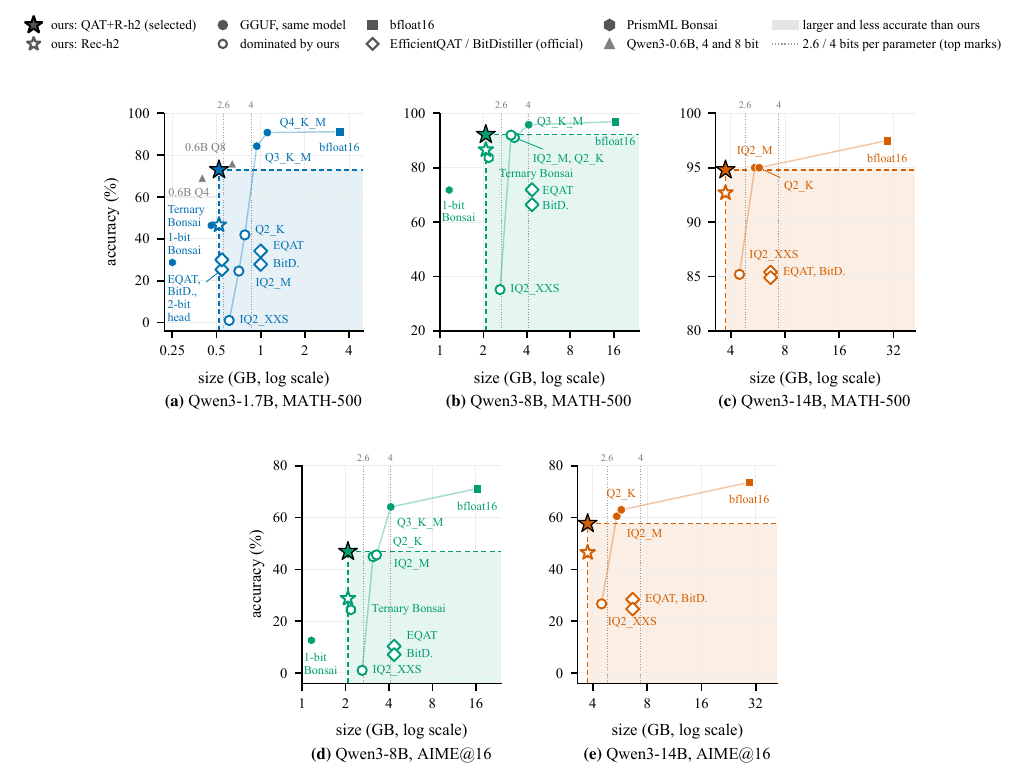}
\caption{Accuracy against whole-model size for every file of each Qwen3 model scored in our harness: \textbf{(a--c)}~sampled, strictly scored MATH-500; \textbf{(d, e)}~AIME@16. Filled stars: \ck{QAT+R-h2} (selected); hollow stars: \ck{Rec-h2}; diamonds: official EfficientQAT and BitDistiller trained on our base mixture (at 1.7B also with a 2-bit head and embedding); hexagons: PrismML Bonsai (ternary and 1-bit); lines join the model's published GGUF files (Unsloth~\citep{unsloth2025qwen3gguf}; UD- prefix omitted) and \texttt{bf16}. The shaded corner holds every file that is both larger and less accurate than the selected checkpoint, i.e.\ Pareto-dominated by it; dominated files are drawn hollow. Grey triangles in (a): Qwen3-0.6B at 4 and 8 bits. Dotted lines: 2.6 and 4 bits per parameter (the size groups of the tables).}
\label{fig:frontierall}
\end{figure}

\paragraph{Larger models lose less.} Two-bit weights cost the most reasoning at small scale. Sampled MATH-500 is 73.0 against \texttt{bf16}'s 91.2 at 1.7B, 92.2 against 96.9 at 8B and 94.8 against 97.5 at 14B, and the 7-task mean is 53.8 against 58.9, 65.8 against 70.7 and 69.9 against 74.3, while the checkpoints take 0.52, 2.08 and 3.74\,GB instead of 3.44, 16.4 and 29.5\,GB. AIME@16, the hardest benchmark, recovers more slowly: 12.3 against 41.4 at 1.7B, 46.9 against 71.2 at 8B and 57.7 against 73.6 at 14B.

\paragraph{Reasoning data for QAT.} Running Stage~4 longer on a mixture with complete reasoning traces (\cref{sec:data}) changes reasoning far more than any stage ablation. Against \ck{Rec-h2} of the same size, sampled MATH-500 rises from 46.7 to 73.0 at 1.7B ($+26.3$ [23.3, 29.4], $p{<}10^{-4}$, paired over problems), from 86.6 to 92.2 at 8B and from 92.7 to 94.8 at 14B, and AIME@16 from 1.0 to 12.3, 28.8 to 46.9 and 46.5 to 57.7. At 1.7B the gain is accuracy on finished traces (83.8 against 54.2) more than finishing them (87.1\% against 86.0\% of traces closed). The 7-task mean moves by at most 2.7 points, while HumanEval falls at 8B and 14B (53.0 to 47.0 and 65.2 to 59.8). On 500 MATH training problems, \ck{QAT+R-h2} scores above \ck{Rec+R-h2}, recalibrated on the same mixture, at every size (63.1 against 45.0, $p{<}10^{-4}$; 88.3 against 86.2, $p{=}0.005$; 92.7 against 91.9, $p{=}0.16$), so the selector (Stage~5) keeps \ck{QAT+R-h2} at 1.7B and 8B; at 14B the two files have the same size and the difference is not significant, and we keep \ck{QAT+R-h2}, the checkpoint with the higher mean.

\paragraph{Against published two-bit models on huggingface.} At every Qwen3 size the selected checkpoint is smaller than the smallest published 2-bit GGUF (UD-IQ2\_XXS~\citep{unsloth-qwen3-1.7b-udiq2xxs,unsloth-qwen3-8b-udiq2xxs,unsloth-qwen3-14b-udiq2xxs}) and more accurate on every task of \cref{tab:scaling}; on sampled MATH-500 it scores 73.0 against 1.0 at 1.7B, 92.2 against 35.2 at 8B and 94.8 against 85.2 at 14B. \Cref{fig:frontierall} shows the resulting Pareto dominance on reasoning tasks. UD-IQ2\_XXS and the official EfficientQAT and BitDistiller files at every size, UD-IQ2\_M~\citep{unsloth-qwen3-1.7b-udiq2m,unsloth-qwen3-8b-udiq2m} and Q2\_K at 1.7B and 8B, and Ternary Bonsai at 8B are all larger and less accurate than our checkpoint. Q2\_K files~\citep{unsloth-qwen3-1.7b-q2k,unsloth-qwen3-8b-q2k,unsloth-qwen3-14b-q2k} (0.78, 3.28 and 5.75\,GB) score 41.9 against our 73.0 at 1.7B ($+31.1$ [27.9, 34.5], $p{<}10^{-4}$), 91.1 against 92.2 at 8B ($+1.1$ [$-$0.5, 2.8], $p{=}0.22$) and 95.0 against 94.8 at 14B ($-$0.2 [$-$1.3, 1.0], $p{=}0.86$): our checkpoints match them at 63--67\% of their size. Files of four or more bits per parameter (Q3\_K\_M and above), and Qwen3-0.6B at eight bits (76.0 at 0.64\,GB), still score higher.

\paragraph{Against two-bit QAT methods and released low-bit models.} The official EfficientQAT and BitDistiller, trained on our base mixture for the same 805M tokens as \ck{Rec-h2}, store a \texttt{bf16} head and embedding and per-group zero points, so their files are 1.8--2.1 times larger than ours. They match or exceed \ck{Rec-h2} on the 7-task mean (53.0 and 51.7 against 51.1 at 1.7B, 66.5 and 65.4 against 64.9 at 8B, 71.2 and 69.8 against 69.9 at 14B) but not on reasoning: on sampled MATH-500 \ck{Rec-h2} is 12.5 and 18.9 points higher at 1.7B (46.7 against 34.2 and 27.8), 14.7 and 20.2 points higher at 8B (86.6 against 71.9 and 66.4; AIME@16 28.8 against 10.3 and 7.1) and 7.3 and 7.8 points higher at 14B (92.7 against 85.4 and 84.9; AIME@16 46.5 against 28.4 and 24.7), all $p{<}10^{-4}$ paired over problems. Quantizing their head and embedding to 2 bits as well (0.54\,GB, byte-matched at 1.7B) lowers MATH-500 to 30.0 and 25.3. PrismML's Bonsai models~\citep{prismml2026bonsai} quantize every linear layer, the embedding and the head. Their training data and pipeline however are not published, so these rows compare released checkpoints. Ternary Bonsai keeps more of the non-reasoning scores (at 0.46\,GB, 7-task 55.0 and HumanEval 48.2 at 1.7B), while \ck{QAT+R-h2} leads on sampled MATH-500 by 26.6 points at 1.7B (73.0 against 46.4) and 8.5 at 8B (92.2 against 83.7; both $p{<}10^{-4}$) and on AIME@16 (12.3 against 1.4, 46.9 against 24.4).

\paragraph{Scaling to 14B.} At Qwen3-14B, the QAT and QAT-only checkpoints (\ck{QAT-h2}, \ck{QAT-only-h2}) are within 0.4 points on the 7-task mean (70.6 against 70.2) and MMLU (70.2 against 69.9); on sampled MATH-500 they score 90.5 and 89.6, and on AIME@16 40.9 and 37.2 (\cref{tab:filesize,tab:scale}). At equal bytes, the recalibrated checkpoint (\ck{Rec-h2}) scores higher than \ck{QAT-h2} on the selector (87.5 against 83.3) and on MATH-500 (92.7 against 90.5). After recalibration, GSQ codewords score higher than layer-wise-start codewords (\ck{Rec-hbf16} against \ck{LW-Rec-hbf16}: 92.8 against 89.7 on MATH-500, 48.2 against 36.7 on AIME@16).

\section{Model size and inference}
\label{sec:storage}
Training checkpoints store codewords in 4-bit containers; we re-pack them at their true widths and decode them again to verify the stored bytes (\cref{app:packed}). With a 2-bit head and embedding, the Qwen3-1.7B QAT and recalibrated checkpoints (\ck{QAT-h2}, \ck{Rec-h2}) each measure 0.520\,GB (2.42 bits per original parameter; 2.046 per quantized weight with scales), and the Qwen3-14B QAT-only and recalibrated checkpoints (\ck{QAT-only-h2}, \ck{Rec-h2}) 3.74\,GB; all four decode bit-identically and load in stock \texttt{transformers}. With a \texttt{bf16} tied embedding the 1.7B checkpoint (\ck{QAT-hbf16}) is 0.977\,GB, 64\% of it embedding. Served by vLLM directly from the 4-bit containers on one H100, the 8B and 14B checkpoints take 5.68 and 9.31\,GiB of weight memory instead of 15.27 and 27.52\,GiB for \texttt{bf16}, and decode 212 and 150 tokens/s at batch 1 instead of 146 and 86 (\cref{app:packed}).

\paragraph{Equal bytes, same harness.}
\Cref{tab:bytes} and \cref{app:frontier} compare our Qwen3-1.7B checkpoints with PrismML's Bonsai releases \citep{prismml2026bonsai}, the official EfficientQAT and BitDistiller trained on our base mixture, and Unsloth's llama.cpp quantizations~\citep{unsloth2025qwen3gguf,gerganov2023llamacpp} of the same model, calibrated with an importance matrix, decoded to \texttt{bf16} and scored with the same harness. At 0.52\,GB, our recalibrated and QAT checkpoints (\ck{Rec-h2}, \ck{QAT-h2}) are smaller than every 2-bit GGUF~\citep{unsloth-qwen3-1.7b-udiq2xxs,unsloth-qwen3-1.7b-udiq2m,unsloth-qwen3-1.7b-q2k}. The recalibrated checkpoint selected under the base mixture (\ck{Rec-h2}), from a second recalibration pass that quantizes the head and embedding, beats Q2\_K~\citep{unsloth-qwen3-1.7b-q2k} on strict sampled MATH-500 (by less under lenient scoring, 47.3 vs.\ 44.8) because more of its traces finish (86\% vs.\ 56\%); on traces both finish, Q2\_K is more accurate, and with a forced answer the two tie (\cref{app:budget}). Our QAT checkpoint (\ck{QAT-h2}) is better on most greedy columns. \ck{QAT+R-h2}, in the same 0.52\,GB, raises sampled MATH-500 to 73.0 and greedy MATH-500 to 68.6, against 41.9 and 27.8 for Q2\_K ($+31.1$ [27.9, 34.5] sampled, $p{<}10^{-4}$), the best of every sub-0.94\,GB Qwen3-1.7B checkpoint; Ternary Bonsai (0.46\,GB) leads the non-reasoning columns but scores 46.4 sampled and 42.8 greedy. The official EfficientQAT and BitDistiller score 34.2 and 27.8 at 1.00\,GB and 30.0 and 25.3 when their head and embedding are also 2-bit (0.54\,GB). It also beats Qwen3-0.6B at 4 bits~\citep{unsloth-qwen3-0.6b-q4km} on MATH-500 (69.2 at 0.40\,GB), but not Qwen3-0.6B at 8 bits~\citep{unsloth-qwen3-0.6b-q80} (76.0 at 0.64\,GB) or Q3\_K\_M~\citep{unsloth-qwen3-1.7b-q3km} (84.3 at 0.94\,GB) (\cref{fig:frontierall}); the 0.6B models trail it on every non-reasoning column.

\begin{table}[!htbp]
\centering
\small
\setlength{\tabcolsep}{3.5pt}
\resizebox{\linewidth}{!}{%
\begin{tabular}{lccccccccc}
\toprule
 & & & & & & & & \multicolumn{2}{c}{MATH-500 (think)} \\
Qwen3-1.7B & GB & bits/param & 7-task & MMLU & GSM8K & HumanEval & BFCL NL & greedy & sampled \\
\midrule
\texttt{bf16} & 3.44 & 16 & 58.9 & 55.5 & 68.6 & 41.5 & 81.7 & 90.2 & 91.2 \\
\midrule
\multicolumn{10}{l}{\emph{$\le$2.6 bits per parameter}} \\
PrismML 1-bit Bonsai$^\circ$ & \textbf{0.25} & \textbf{1.15} & 49.7 & 41.7 & 43.9 & 37.2 & 74.3 & 25.4 & 28.7 \\
PrismML Ternary Bonsai$^\circ$ & 0.46 & 2.15 & \textbf{55.0} & \textbf{48.6} & \textbf{60.0} & \textbf{48.2} & \textbf{74.7} & 42.8 & 46.4 \\
\rowcolor{ourrow}Ours: \ck{QAT+R-h2} (selected)$^\S$ & 0.52 & 2.42 & 53.8 & 47.0 & 58.7 & 25.6 & 73.0 & \textbf{68.6} & \textbf{73.0} \\
\rowcolor{ourrow}Ours: \ck{Rec-h2} & 0.52$^\ddagger$ & 2.42 & 51.1 & 45.2 & 51.8 & 22.6 & 73.0 & 39.0 & 46.7 \\
\rowcolor{ourrow}Ours: \ck{QAT-h2} & 0.52 & 2.42 & 54.0 & 48.0 & 54.1 & 20.1 & 72.0 & 38.8 & 42.0 \\
EfficientQAT (official), 2-bit head/emb.$^\dagger$ & 0.54 & 2.53 & 51.9 & 45.8 & 43.7 & 22.0 & 58.4 & 31.6 & 30.0 \\
BitDistiller (official), 2-bit head/emb.$^\dagger$ & 0.54 & 2.53 & 51.0 & 45.6 & 42.4 & 16.5 & 61.4 & 21.6 & 25.3 \\
\multicolumn{10}{l}{\emph{2.6--4 bits per parameter}} \\
GGUF UD-IQ2\_XXS~\citep{unsloth-qwen3-1.7b-udiq2xxs} & 0.61 & 2.82 & 38.2 & 26.0 & 1.3 & 1.2 & 0.0 & 2.8 & 1.0 \\
GGUF UD-IQ2\_M~\citep{unsloth-qwen3-1.7b-udiq2m} & 0.71 & 3.30 & 48.6 & 46.4 & 22.9 & 15.2 & 42.8 & 14.2 & 24.6 \\
GGUF Q2\_K~\citep{unsloth-qwen3-1.7b-q2k} & 0.78 & 3.62 & 49.2 & 40.0 & 31.8 & 18.3 & 54.5 & 27.8 & 41.9 \\
\multicolumn{10}{l}{\emph{$>$4 bits per parameter}} \\
GGUF Q3\_K\_M~\citep{unsloth-qwen3-1.7b-q3km} & 0.94 & 4.37 & 55.7 & 53.1 & 60.9 & 26.2 & 80.6 & 81.2 & 84.3 \\
\rowcolor{ourrow}Ours: \ck{QAT-hbf16} & 0.98 & 4.54 & 54.6 & 49.9 & 50.2 & 17.7 & 75.5 & 38.2 & 43.0 \\
\rowcolor{ourrow}Ours: \ck{Rec-hbf16} & 0.98$^\ast$ & 4.54 & 51.5 & 44.4 & 51.6 & 22.6 & 72.2 & 40.6 & 47.5 \\
EfficientQAT (official)$^\dagger$ & 1.00 & 4.65 & 53.0 & 46.2 & 42.3 & 22.6 & 62.3 & 33.2 & 34.2 \\
BitDistiller (official)$^\dagger$ & 1.00 & 4.65 & 51.7 & 45.9 & 41.4 & 20.7 & 63.8 & 25.8 & 27.8 \\
GGUF Q4\_K\_M~\citep{unsloth-qwen3-1.7b-q4km} & 1.11 & 5.15 & 58.7 & 55.7 & 66.0 & 33.5 & 82.3 & 88.4 & 90.8 \\
\midrule
\multicolumn{10}{l}{\emph{On Qwen3-0.6B}} \\
Qwen3-0.6B GGUF Q4\_K\_M~\citep{unsloth-qwen3-0.6b-q4km} & 0.40 & 5.4$^\P$ & 46.5 & 35.0 & 33.9 & 17.7 & 70.4 & 61.8 & 69.2 \\
Qwen3-0.6B GGUF Q8\_0~\citep{unsloth-qwen3-0.6b-q80} & 0.64 & 8.6$^\P$ & 48.7 & 40.1 & 41.7 & 20.1 & 72.9 & 69.2 & 76.0 \\
\bottomrule
\end{tabular}}
\caption{At 0.52\,GB, \ck{QAT+R-h2} reasons better than every Qwen3-1.7B file of up to 4 bits per parameter; Ternary Bonsai leads the non-reasoning columns. Same harness for all files; bits/param use the original 1.72B parameters. Rows grouped by bits per parameter; bold: best in the $\le$2.6-bit group. $^\S$QAT on the reasoning mixture. $^\ddagger$Second recalibration pass that quantizes the head and embedding. $^\ast$Not separately re-packed. $^\dagger$Official code on our base mixture, same tokens. $^\circ$Released files \citep{prismml2026bonsai}. $^\P$Bits per Qwen3-0.6B parameter. BFCL NL: BFCL non-live.}
\label{tab:bytes}
\end{table}

\section{Controls behind each stage}
\label{sec:controls}
\begin{figure}[!htbp]
\centering
\newsavebox{\recipebox}%
\begin{lrbox}{\recipebox}\small\renewcommand{\arraystretch}{1.55}%
\begin{tabular}{@{}c@{\hspace{7pt}}p{0.2\linewidth}@{\hspace{6pt}}p{0.57\linewidth}@{\hspace{6pt}}r@{}}
\multicolumn{4}{@{}l}{\color{stA}\footnotesize\itshape Post-training quantization, block by block} \\[-3pt]
\rowcolor{stAbg}\stagebadge{stA}{1} & \textbf{Allocate bits} & KronQ sensitivity sets each layer to ternary, 2 or 3 bits & 2.0 bpw \\
\rowcolor{stAbg}\stagebadge{stA}{2} & \textbf{Search codewords} & GSQ from a layer-wise start, matching \texttt{bf16} block outputs &  \\
\multicolumn{4}{@{}l}{\color{stB}\footnotesize\itshape Whole-model training, distilled from \texttt{bf16}} \\[-3pt]
\rowcolor{stBbg}\stagebadge{stB}{3} & \textbf{Recalibrate scales} & Codewords frozen; scales and RMSNorm gains fit by KL &  \\
\rowcolor{stBbg}\stagebadge{stC}{4} & \textbf{Reassign codewords} & QAT from a residual start ($\alpha{=}0.5$) &  \\
\multicolumn{4}{@{}l}{\color{stD}\footnotesize\itshape Selection} \\[-3pt]
\rowcolor{stDbg}\stagebadge{stD}{5} & \textbf{Select checkpoint} & Paired test on MATH training problems keeps the better of 3 and 4, or the smaller on a tie &  \\
\end{tabular}\end{lrbox}%
\begin{tikzpicture}
\node[draw=boxedge, line width=2pt, rounded corners=6pt, fill=white, inner xsep=7pt, inner ysep=9pt] (body) {\usebox{\recipebox}};
\node[fill=boxedge, text=white, rounded corners=3pt, font=\small\bfseries, inner xsep=8pt, inner ysep=3.5pt] at (body.north) {Two-bit scalar quantization for reasoning models};
\end{tikzpicture}
\caption{The framework at a glance. Every weight is an integer codeword times a \texttt{bf16} scale shared by a group of 128 weights; one checkpoint is kept per model. The rest of this section gives the evidence for each stage.}
\label{fig:recipe}
\end{figure}
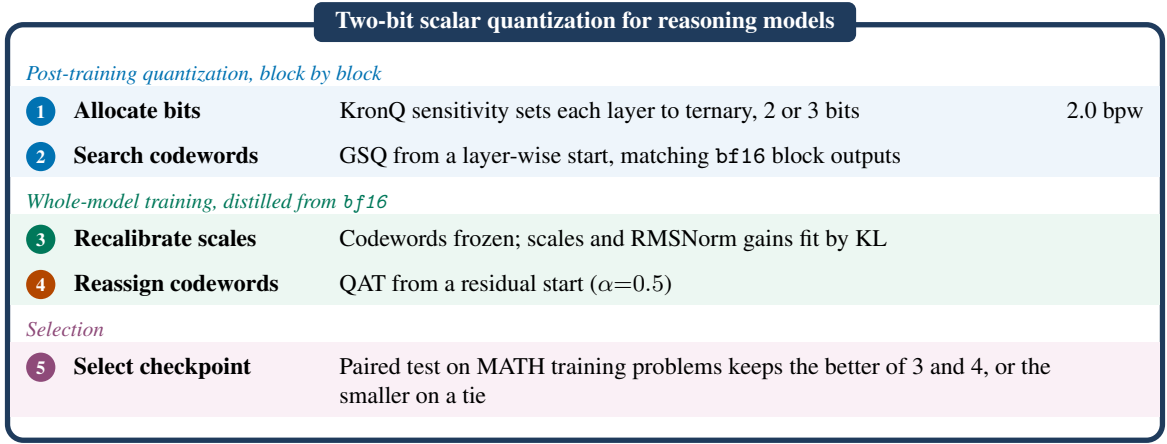

\Cref{fig:recipe} summarizes \texttt{ThinkBit}. Each control below removes one stage or design choice under the same frozen hyper-parameters and evaluation (\cref{tab:controls}); \cref{tab:reasoning} gives the paired reasoning contrasts. Recalibration plus QAT uses 805M tokens and the no-recalibration QAT arms 719M; held-out KL is still falling at the last step.

\begin{table}[!htbp]
\centering
\small
\setlength{\tabcolsep}{3.5pt}
\resizebox{\linewidth}{!}{%
\begin{tabular}{lccccccc}
\toprule
 & & & & & & \multicolumn{2}{c}{MATH-500 (thinking)} \\
\cmidrule(l){7-8}
Qwen3-1.7B variant & WikiText-2 $\downarrow$ & 7-task $\uparrow$ & MMLU $\uparrow$ & GSM8K $\uparrow$ & HumanEval $\uparrow$ & greedy $\uparrow$ & sampled $\uparrow$ \\
\midrule
\texttt{bf16} & 16.67 & 58.9 & 55.5 & 68.6 & 41.5 & 90.2 & 91.2 \\
\ck{GSQ-hbf16} & 46.91 & 43.8 & 36.4 & 27.5 & 8.5 & 15.0 & 14.8 \\
\midrule
\multicolumn{8}{l}{\emph{Scale recalibration (codewords frozen)}} \\
\quad RMSNorm gains only & 25.29 & 48.0 & 38.2 & 38.8 & 16.5 & 27.8 & 34.0 \\
\quad scales only & 19.49 & 51.4 & 44.1 & 51.3 & 23.2 & 39.2 & \textbf{48.5} \\
\quad scales + RMSNorm gains (\ck{Rec-hbf16}) & 19.46 & 51.5 & 44.4 & 51.6 & 22.6 & \textbf{40.6} & 47.5 \\
\midrule
\multicolumn{8}{l}{\emph{Instead of QAT: re-round toward \texttt{bf16}, then briefly recalibrate scales}} \\
\quad codewords with $|W/s-c|>0.5$ that can step (23\%) & 26.51 & 44.5 & 35.0 & 13.0 & 7.9 & 6.4 & 5.5 \\
\quad 6.4\% with the largest movable $|W/s-c|$ & 20.23 & 49.6 & 41.2 & 43.6 & 14.6 & 32.2 & 35.4 \\
\midrule
\multicolumn{8}{l}{\emph{After QAT}} \\
\rowcolor{ourrow}\quad \textbf{\ck{QAT-hbf16}: full framework (reference)} & 16.54 & 54.6 & \textbf{49.9} & 50.2 & 17.7 & 39.0 & 43.0 \\
\quad \ck{LW-QAT-hbf16}: layer-wise start instead of GSQ & \textbf{16.53} & 53.7 & 48.7 & 47.9 & 24.4 & 33.0 & 35.0 \\
\quad \ck{QAT-noRec-hbf16}: without the scale recalibration (719M tokens) & 16.68 & 54.5 & 48.9 & 52.2 & 26.2 & 34.2 & 37.0 \\
\quad \ck{QAT-only-hbf16}: layer-wise start, without recalibration & \textbf{16.53} & \textbf{54.8} & \textbf{49.9} & 46.9 & 18.3 & 26.6 & 32.1 \\
\quad \quad \ck{QAT-only-h2}: + untied 2-bit head/emb. & 17.44$^\circ$ & \textbf{54.8} & 48.5 & 47.2 & 15.2 & 26.6 & 29.1 \\
\quad \quad codewords frozen ($\alpha{=}0$) & 16.99$^\circ$ & 53.6 & 48.7 & 46.9 & 14.0 & 24.4 & 30.1 \\
\quad uniform 2-bit (2.125 bpw), GPTQ, recalibration & 16.74 & 54.4 & 49.2 & 45.8 & 18.3 & 25.4 & 28.4 \\
\quad without Wikipedia and MMLU-aux data & 17.23 & 54.2 & 47.8 & \textbf{56.0} & 15.9 & 36.2 & 42.5 \\
\quad also without Evol-Instruct code & 16.98 & 54.3 & 48.7 & 55.1 & 18.9 & 37.0 & 41.6 \\
\quad revert MLP codeword reassignments & 18.99$^\circ$ & 51.3 & 43.1 & 47.8 & \textbf{29.9} & 39.6 & 42.1 \\
\quad revert attention codeword reassignments & 18.47$^\circ$ & 53.5 & 48.9 & 45.9 & 12.2 & 34.6 & 39.4 \\
\quad revert value-projection reassignments & 17.48$^\circ$ & 54.7 & \textbf{49.9} & 46.5 & 17.1 & 35.6 & 38.8 \\
\bottomrule
\end{tabular}}
\caption{ Removing any stage keeps the 7-task mean within 1.5 points but costs 6.0--14.6 points of sampled MATH-500. Qwen3-1.7B variants (\texttt{bf16} head and embedding); paired tests in \cref{tab:reasoning}. Shaded: full framework; bold: best quantized value. ``Revert'' restores the pre-QAT codewords in one layer class. $^\circ$WikiText-2 from a separate pass (\texttt{bf16}: 16.72).}
\label{tab:controls}
\end{table}

\begin{table}[!htbp]
\centering
\small
\setlength{\tabcolsep}{4pt}
\resizebox{\linewidth}{!}{%
\begin{tabular}{llccccc}
\toprule
Model & Contrast (A vs.\ B) & A & B & $\Delta$ [95\% CI] & $p$ & status \\
\midrule
\rowcolor{ourrow}Qwen3-1.7B & codeword search (\ck{QAT} vs.\ \ck{LW-QAT}) & 43.0 & 35.0 & \textbf{+8.0} [5.6, 10.2] & $<10^{-4}$ & abl \\
 & \quad vs.\ tuned GPTQ start (all tiers, clipped scales) & 43.0 & 32.6 & \textbf{+10.3} [8.0, 12.8] & $<10^{-4}$ & check \\
 & \quad vs.\ GPTQ start + Hessian-weighted rounding & 43.0 & 37.8 & \textbf{+5.2} [3.1, 7.5] & $<10^{-4}$ & check \\
\rowcolor{ourrow} & recalibration before QAT (\ck{QAT} vs.\ \ck{QAT-noRec}) & 43.0 & 37.0 & \textbf{+6.0} [3.6, 8.3] & $<10^{-4}$ & abl \\
 & recalibration on layer-wise-start codewords (\ck{LW-QAT} vs.\ \ck{QAT-only}) & 35.0 & 32.1 & +2.85 [0.85, 4.90] & 0.009 & abl \\
\rowcolor{ourrow} & both (\ck{QAT} vs.\ \ck{QAT-only}) & 43.0 & 32.1 & \textbf{+10.8} [8.5, 13.0] & $<10^{-4}$ & abl \\
 & QAT stage (\ck{QAT} vs.\ \ck{Rec}) & 43.0 & 47.5 & \textbf{$-$4.5} [$-$7.0, $-$2.1] & $8\cdot10^{-4}$ & abl \\
\rowcolor{ourrow} & same-data QAT baseline (\ck{QAT} vs.\ EfficientQAT port, $\approx$1.0\,GB) & 43.0 & 28.8 & \textbf{+14.2} [11.8, 16.8] & $<10^{-4}$ & check \\
 & codeword search at the recalibrated checkpoint (\ck{Rec} vs.\ \ck{LW-Rec})$^\S$ & 47.2 & 22.8 & \textbf{+24.4} [21.8, 27.2] & $<10^{-4}$ & check \\
 & \quad vs.\ tuned GPTQ start (all tiers, clipped scales) & 47.5 & 18.5 & \textbf{+28.9} [25.9, 32.1] & $<10^{-4}$ & check \\
 & \quad vs.\ GPTQ start + Hessian-weighted rounding & 47.5 & 24.1 & \textbf{+23.4} [20.4, 26.4] & $<10^{-4}$ & check \\
\midrule
\multicolumn{7}{l}{\emph{Checkpoint selection: 313 MATH training problems}} \\
Qwen3-1.7B & \ck{QAT-hbf16} vs.\ \ck{Rec-hbf16} (0.98\,GB) & 22.5 & 28.8 & \textbf{$-$6.2} [$-$9.0, $-$3.5] & $10^{-4}$ & sel \\
Qwen3-1.7B & \ck{QAT-h2} vs.\ \ck{Rec-h2} (0.52\,GB) & 21.9 & 27.6 & \textbf{$-$5.7} [$-$7.9, $-$3.4] & $<10^{-4}$ & sel \\
Qwen3-1.7B & \ck{Rec-h2} vs.\ \ck{Rec-hbf16} (0.52 vs.\ 0.98\,GB) & 27.6 & 28.8 & $-$1.2 [$-$3.7, 1.1] & 0.37 & sel \\
Qwen3-1.7B & codeword search at the recalibrated checkpoint (\ck{Rec} vs.\ \ck{LW-Rec})$^\S$ & 27.8 & 8.6 & \textbf{+19.2} [15.9, 22.7] & $<10^{-4}$ & sel \\
Qwen3-14B & \ck{QAT-h2} (3.74\,GB) vs.\ \ck{Rec-hbf16} (6.4\,GB) & 83.3 & 87.9 & \textbf{$-$4.6} [$-$6.6, $-$2.7] & $<10^{-4}$ & sel \\
 & \ck{QAT-h2} vs.\ \ck{Rec-h2} (3.74\,GB) & 83.3 & 87.5 & \textbf{$-$4.2} [$-$6.3, $-$2.1] & $2\cdot10^{-4}$ & sel \\
\bottomrule
\end{tabular}}
\caption{Codeword search and recalibration each raise sampled MATH-500 significantly, and every selector decision agrees with MATH-500. Paired contrasts on MATH-500 ($n{=}500$) and on the 313 selection problems. Bold: significant after Holm correction. Status: abl, stage ablation; check, robustness check; sel, selector decision. $^\S$Separate evaluation run (\ck{Rec}: 47.2 there, 47.45 in the main run).}
\label{tab:reasoning}
\end{table}

\paragraph{Reasoning performance differentiates between the configurations.} After QAT, removing codeword search, scale recalibration, both, or mixed bit placement keeps the 7-task mean at 53.7--54.8, MMLU at 48.7--49.9 and WikiText-2 at 16.53--16.74 (full framework: 54.6, 49.9 and 16.54), while sampled MATH-500 falls from 43.0 to 28.4--37.0 (\cref{tab:controls}). With a pre-stated $\pm$1.5-point margin, every removal is equivalent on the 7-task mean; MMLU differences of up to 2.1 points are not equivalent within the margin in 4 of 6 contrasts (\cref{app:morecontrols}). QAT itself raises the 7-task mean from 51.5 to 54.6 but lowers sampled MATH-500 from 47.5 to 43.0. HumanEval does not follow the same ordering, so the finding concerns long-form reasoning. A rule with a log-likelihood conjunct would have dropped the search (\cref{app:morecontrols}).

\paragraph{QAT baselines.} We apply EfficientQAT~\citep{chen2024efficientqat} trained on our data for the same 805M tokens (1.00\,GB with a \texttt{bf16} head) keeps a 7-task mean of 52.0 but scores 28.8 on sampled MATH-500, against 43.0 for the QAT checkpoint of about the same size (\ck{QAT-hbf16}) and 47.5 for the recalibrated checkpoint (\ck{Rec-hbf16}) (both $p{<}10^{-4}$); it closes 56\% of its traces (\cref{tab:decompall,app:morecontrols}). The official EfficientQAT code on the same data scores 34.2 (\cref{tab:bytes}). Furthermore at every model size EfficientQAT outperforms BitDistiller \citep{du2024bitdistiller} on reasoning (MATH-500 and AIME@16).

\paragraph{Checkpoint selection without the test set.} The framework produces two candidates per model, the recalibrated and the QAT checkpoint (\ck{Rec}, \ck{QAT}), and the selector decides which one to keep. It samples four completions for each of 313 level-3--5 MATH training problems (disjoint from MATH-500), scores strictly, and keeps the higher mean if the paired sign-flip test gives $p\le0.05$, else the smaller checkpoint. At 1.7B it keeps the recalibrated checkpoint at both 0.98 and 0.52\,GB (\ck{Rec-hbf16}, \ck{Rec-h2}), and the smaller of the two; at 14B it keeps the 3.74\,GB recalibrated checkpoint (\ck{Rec-h2}), which beats the QAT checkpoint of equal size (\ck{QAT-h2}) and ties the 6.4\,GB \texttt{bf16}-head version (\ck{Rec-hbf16}) (\cref{app:morecontrols}). MATH-500 agrees with every decision (\cref{tab:filesize,tab:selaudit}). With the reasoning mixture, the selector compares \ck{QAT+R-h2} with \ck{Rec+R-h2} (\cref{sec:scaling}). At 0.52\,GB the preference is a termination effect: forced answers tie (\cref{app:budget}).

\section{Related work}
\label{sec:related}

\paragraph{Post-training quantization of LLMs.} The first LLM quantizers targeted 8-bit weights and activations, using fine-grained quantization with layer-wise distillation (ZeroQuant~\citep{yao2022zeroquant}), mixed-precision decomposition of outlier features (LLM.int8()~\citep{dettmers2022llmint8}) or rescaling that moves outliers from activations to weights (SmoothQuant~\citep{xiao2023smoothquant}). For weight-only quantization, GPTQ~\citep{frantar2023gptq} turns Optimal Brain Compression~\citep{frantar2022obc} into one-shot second-order rounding, AWQ~\citep{lin2024awq} protects salient channels by rescaling, and SpQR~\citep{dettmers2024spqr} and SqueezeLLM~\citep{kim2024squeezellm} store sparse outlier weights at higher precision. Learned rounding and clipping optimize layer or block reconstruction (AdaRound~\citep{nagel2020adaround}, BRECQ~\citep{li2021brecq}, OmniQuant~\citep{shao2024omniquant}, CBQ~\citep{ding2025cbq}, TesseraQ~\citep{li2024tesseraq}). Most of these methods are close to lossless at three to four bits but degrade sharply at two. GSQ~\citep{dadgarnia2026gsq}, which we build on, learns each weight's grid assignment and the group scales jointly through a Gumbel-softmax relaxation of a scalar grid.

\paragraph{Two-bit and sub-two-bit formats.} The strongest two-bit results leave the scalar format. Incoherence processing (QuIP~\citep{chee2023quip}), combined with lattice or trellis codebooks (QuIP\#~\citep{tseng2024quipsharp}, QTIP~\citep{tseng2024qtip}), and additive vector quantization with fine-tuning (AQLM~\citep{egiazarian2024aqlm}, PV-Tuning~\citep{malinovskii2024pvtuning}) reach strong perplexity, and fixed or learned rotations (QuaRot~\citep{ashkboos2024quarot}, SpinQuant~\citep{liu2025spinquant}) remove outliers so that weights, activations and the KV cache are easier to quantize. Binary and partially binarized quantization (BiLLM~\citep{huang2024billm}, PB-LLM~\citep{shang2024pbllm}) reaches about one bit per weight at a clear accuracy cost, while binary and ternary models trained from scratch (BitNet~\citep{wang2023bitnet}, BitNet b1.58~\citep{ma2024bitnet158}, Spectra~\citep{kaushal2024spectra}) show that very low-bit weights can be competitive when the pretraining budget allows. We keep a strictly scalar format with one \texttt{bf16} scale per group of 128 weights, which dequantizes with one multiplication and needs no codebook or rotation at inference; rotations are orthogonal to our stages and could be combined with them.

\paragraph{Quantization-aware training and distillation.} QAT trains through the rounding with straight-through estimators~\citep{bengio2013ste,jacob2018quantization} or learned step sizes~\citep{esser2020lsq}. For LLMs, LLM-QAT~\citep{liu2023llmqat} distills from the full-precision model on self-generated data, BitDistiller~\citep{du2024bitdistiller} combines asymmetric quantization with self-distillation, EfficientQAT~\citep{chen2024efficientqat} trains all parameters block by block and then only the quantization step sizes end to end, OneBit~\citep{xu2024onebit} factors weights into signs and value vectors, and ParetoQ~\citep{liu2025paretoq} compares 1- to 4-bit QAT within one framework. Parameter-efficient methods train only scales, adapters or normalization layers on frozen low-bit weights (PEQA~\citep{kim2023peqa}, QLoRA~\citep{dettmers2023qlora}, norm tweaking~\citep{li2024normtweaking}); our scale recalibration is closest to these, but fits scales and gains to the \texttt{bf16} model's output distribution with codewords frozen. Our recalibration and QAT distill with the forward KL to the \texttt{bf16} model~\citep{hinton2015distilling}; reverse-KL and on-policy distillation~\citep{gu2024minillm,agarwal2024gkd} are alternatives we do not use. We show that, at a fixed token budget, the post-training starting point and a reasoning-based selector decide how much reasoning survives.

\paragraph{Mixed precision and sensitivity.} Since HAWQ, second-order sensitivity has guided bit allocation~\citep{dong2019hawq,dong2020hawqv2}. KronQ~\citep{kronq} scores each layer with the activation and gradient traces of a Kronecker-factored Hessian approximation (as in K-FAC~\citep{martens2015kfac}); we use its score to choose ternary, 2-bit or 3-bit tiers under a bit budget (\cref{sec:method}).

\paragraph{Evaluating quantized models.} Systematic evaluations find that perplexity can miss losses on knowledge-intensive tasks~\citep{jaiswal2024llmkick} and that quantization affects dialogue and long-context tasks differently from basic NLP tasks~\citep{li2024evaluating}, although perplexity remains a reasonable proxy on many standard benchmarks~\citep{jin2024comprehensive}; perplexity is also unreliable for long-context ability in general~\citep{fang2025longppl}. Matched accuracy can hide answer flips~\citep{dutta2024accuracy}, and 8-bit and 4-bit weight-only formats are close to lossless across academic and real-world tasks~\citep{kurtic2024bf16}. Scaling studies relate the useful precision to model size and training tokens~\citep{kumar2025scaling,ouyang2024lowbit}. For reasoning models, quantization below four bits hurts most on hard problems~\citep{liu2025quanthurts}; whether it lengthens traces is disputed, with no increase reported by \citet{liu2025quanthurts} and token inflation at 3 and 4 bits by \citet{lian2026inflates}. Qwen3 has been evaluated down to 1 bit with perplexity and zero-shot accuracy~\citep{zheng2025qwen3quant}; reasoning-specific quantizers use fine-tuning signals~\citep{zhang2026quantlrm} or the model's own reasoning traces as calibration data~\citep{wang2026attend}; and budget forcing~\citep{muennighoff2025s1} controls how long a model thinks. These studies compare a quantized model with its full-precision parent. We use paired tests~\citep{miller2024errorbars} to show that likelihood also fails to rank competing two-bit configurations, and select checkpoints on held-out reasoning problems instead.

\section{Conclusion}
\label{sec:conclusion}
We propose \texttt{ThinkBit}, a strictly scalar two-bit quantization framework designed to preserve reasoning. \texttt{ThinkBit} allocates a ternary, 2-bit or 3-bit tier to each layer by sensitivity, searches the integer codewords block by block, recalibrates the stored scales with the codewords frozen, lets quantization-aware training reassign codewords, and keeps the better checkpoint by a paired reasoning test on training problems. Our main finding is that likelihood-based metrics hide what two-bit quantization does to reasoning: removing codeword search, scale recalibration or tier allocation barely changes the log-likelihood mean but clearly lowers sampled MATH-500, and selecting checkpoints by a reasoning test, without touching the test set, picks the stronger model. Trained on the same data, our checkpoints reason better than EfficientQAT and BitDistiller at every size we compared. With reasoning traces in the final training stage, the selected checkpoints are smaller than every published 2-bit GGUF of the same model; they reason better than all of them at 1.7B, match Q2\_K at 8B and 14B at about two thirds of its size, and at 14B come within a few points of \texttt{bf16} on MATH-500. Files of four or more bits per parameter remain stronger, and some two-bit reasoning traces still do not finish.

There are many potential directions for future work. First, kernels that read 2- and 3-bit codewords directly would turn the smaller files into proportional memory and speed gains at inference time, which the 4-bit containers used here cannot show. Second, training objectives that reward finishing a reasoning trace, such as on-policy distillation from the \texttt{bf16} teacher or reinforcement learning after quantization, may close much of the remaining gap to \texttt{bf16}. Third, the stage study should be repeated on more model families, mixture-of-experts models and models above 14B, and the tier allocation could be extended below two bits. Finally, selectors for code and tool use would extend checkpoint selection beyond math problems. We will release the code upon acceptance.

\FloatBarrier
\bibliographystyle{plainnat}
\bibliography{references}

\clearpage
\appendix

\section{Full algorithmic procedure and notation}
\label{app:algorithm}
\Cref{alg:full} presents \texttt{ThinkBit} as procedures, from the \texttt{bf16} model to the selected checkpoint; \cref{tab:notation} collects the notation of \cref{sec:method}.

\begin{algorithm}[H]
\caption{\texttt{ThinkBit}, end to end}
\label{alg:full}
\small
\begin{algorithmic}[1]
\Function{TwoBit}{$M_{\texttt{bf16}}$, $\mathcal{D}_{\mathrm{cal}}$, $\mathcal{D}_{\mathrm{train}}$, $\mathcal{P}$, $\bar B$}
  \State $S\gets\Call{Sensitivity}{M_{\texttt{bf16}},\mathcal{D}_{\mathrm{cal}}}$ \Comment{Stage 1, \cref{eq:sensitivity}}
  \State $b\gets\Call{Allocate}{S,\bar B}$ \Comment{\cref{alg:alloc}}
  \State $(c,s)\gets\Call{CodewordSearch}{M_{\texttt{bf16}},b,\mathcal{D}_{\mathrm{cal}}}$ \Comment{Stage 2}
  \State $M_{\mathrm{rec}}\gets\Call{Recalibrate}{c,s,\mathcal{D}_{\mathrm{train}}}$ \Comment{Stage 3}
  \State $M_{\mathrm{QAT}}\gets\Call{QAT}{M_{\mathrm{rec}},\mathcal{D}_{\mathrm{train}}}$ \Comment{Stage 4}
  \State \Return $\Call{Select}{M_{\mathrm{rec}},M_{\mathrm{QAT}},\mathcal{P}}$ \Comment{Stage 5}
\EndFunction
\Statex
\Function{CodewordSearch}{$M$, $b$, $\mathcal{D}$}
  \State $X\gets$ inputs of block 1 on $\mathcal{D}$
  \For{block $k=1,\dots,N_{\mathrm{blk}}$}
    \For{group $g\in$ (attention, MLP) of block $k$} \Comment{$H_X$ from the group's current inputs}
      \For{layer $\ell$ in $g$} \Comment{LayerRefine: \cref{eq:gptq} via \cref{eq:gumbel,eq:ternary,eq:reinmax}}
        \If{$b_\ell$ is ternary}
          \State $s_\ell\gets$ GPTQ's scale rule on $W_\ell$;\ \ $(c^0_\ell,s_\ell)\gets\Call{LayerRefine}{\text{uninformative logits},s_\ell}$
        \Else
          \State $(c^0_\ell,s_\ell)\gets\Call{GPTQ}{W_\ell,\,H_X+\mu I,\,b_\ell}$ \Comment{\cref{eq:gptq,eq:damping}}
          \State \textbf{if} $\ell$ is a query or key projection \textbf{then} $(c^0_\ell,s_\ell)\gets\Call{LayerRefine}{c^0_\ell,s_\ell}$
        \EndIf
      \EndFor
      \State $z\gets\Call{InitLogits}{c^0}$ for every layer of $g$ except query and key, which keep $c^0$
      \State $(z,s)\gets\arg\min\ \mathcal{L}_{\mathrm{blk}}(z,s;X)$, annealing $\tau,\kappa$ \Comment{\cref{eq:block}}
      \State $c\gets$ the most likely value of each codeword
    \EndFor
    \State $X\gets\mathrm{Block}^{q}_k(X)$ \Comment{inputs of the next block}
  \EndFor
  \State \Return $(c,s)$
\EndFunction
\Statex
\Function{Recalibrate}{$c$, $s_0$, $\mathcal{D}$} \Comment{$c$ frozen}
  \State $(r,\psi)\gets\arg\min_{r,\psi}\mathcal{L}_{\mathrm{rec}}$ on $\mathcal{D}$ with $r,\psi$ clamped, best held-out KL \Comment{\cref{eq:recal}}
  \State \Return model with $s=s_0e^{r}$, $\nu=\nu_0(1+\psi)$
\EndFunction
\Statex
\Function{QAT}{$M$, $\mathcal{D}$}
  \State $u\gets c+\operatorname{clip}\bigl(\alpha(W/s-c),\pm\beta\bigr)$ \Comment{\cref{eq:residual}}
  \State $(u,r,\psi)\gets\arg\min\mathcal{L}_{\mathrm{QAT}}$ with the STE, scales and gains still training, best held-out KL \Comment{\cref{eq:ste,eq:qatloss}}
  \State \Return model with $c=\operatorname{round}(\bar u)$ and the trained scales and gains
\EndFunction
\Statex
\Function{Select}{$A$, $B$, $\mathcal{P}$}
  \State $\mathrm{acc}^A_i,\mathrm{acc}^B_i\gets$ fraction of 4 sampled answers that are strictly correct, for each $i\in\mathcal{P}$
  \State $p\gets\Call{SignFlip}{\mathrm{acc}^A-\mathrm{acc}^B}$ \Comment{\cref{eq:signflip}}
  \If{$p\le0.05$} \Return the candidate with higher mean accuracy \Else\ \Return the smaller candidate (the recalibrated one if equal in size) \EndIf
\EndFunction
\end{algorithmic}
\end{algorithm}

\begin{table}[H]
\centering
\small
\setlength{\tabcolsep}{4pt}
\begin{tabular}{L{3.2cm}L{9.6cm}}
\toprule
Symbol & Meaning \\
\midrule
\multicolumn{2}{l}{\emph{\Cref{alg:full}}} \\
$M$, $\mathcal{D}_{\mathrm{cal}}$, $\mathcal{D}_{\mathrm{train}}$, $\mathcal{P}$ & model; calibration and training sequences; selection problems \\
$N_{\mathrm{blk}}$, $X$ & number of transformer blocks; calibration activations entering the current block \\
\multicolumn{2}{l}{\emph{Format (\cref{sec:format})}} \\
$W_\ell$, $\widehat W_\ell$, $n_\ell$ & \texttt{bf16} and quantized weight of linear layer $\ell$ ($d_{\mathrm{out}}\times d_{\mathrm{in}}$); number of weights \\
$G$, $\gamma(j)$ & group size (128); group of input column $j$ \\
$c_{ij}$, $s_{i,\gamma(j)}$ & integer codeword and \texttt{bf16} group scale of weight $(i,j)$ \\
$b$, $\mathcal{C}_b$, $\mathcal{T}_\ell$ & tier (ternary, 2-bit, 3-bit); its grid; tiers allowed for layer $\ell$ \\
$B_b$, $\bar B$ & bits per weight of tier $b$ (\cref{eq:bits}); budget in decoder bits per weight \\
\multicolumn{2}{l}{\emph{Stage 1: bit allocation (\cref{sec:stage1})}} \\
$\xi_t$, $p_\theta$ & token $t$ of a sequence; \texttt{bf16} next-token distribution \\
$x_t$, $y_t$, $g_t$ & layer input, layer output, and loss gradient at the output, at token $t$ \\
$H_X$, $H_G$ & second moments of $x_t$ and $g_t$ (Kronecker factors of the Hessian) \\
$E_\ell$, $\sigma_{E,\ell}^2$ & rounding error $\widehat W_\ell-W_\ell$; its per-entry variance \\
$S_\ell$, $N_b$ & sensitivity of layer $\ell$ (\cref{eq:sensitivity}); rounding noise of tier $b$ (\cref{eq:noise}) \\
$\zeta_j$, $Q_b$ & standard normal weights; rounding to the nearest point of $\mathcal{C}_b$ \\
$\lambda$, $\mathrm{bpw}(\lambda)$, $b_\ell(\lambda)$ & price per bit; resulting average bits per weight; tier chosen by layer $\ell$ \\
\multicolumn{2}{l}{\emph{Stage 2: codeword search (\cref{sec:stage2})}} \\
$H$, $\mu$, $\mu_0$, $\phi_o$ & damped GPTQ Hessian; damping and its base factor; second moment of output channel $o$ \\
$R$, $e_j$, $\rho$ & upper Cholesky factor of $H^{-1}$; scaled error of column $j$; scale shrink factor \\
$c^0$, $v_k$, $\delta_k$, $z$ & starting codeword; candidate values; their offsets from $c^0$; logits \\
$\epsilon_k$, $U_k$, $\eta$ & Gumbel noise; its uniform source; logistic noise \\
$\pi$, $\tilde c$ & relaxed one-hot vector; soft codeword \\
$\tau$, $\kappa$ & temperature; logit sharpening factor \\
$m$, $\sigma$, $z^{m}$, $z^{\sigma}$ & ternary mask and sign; their logits \\
$q$, $\omega$, $J(\pi)$, $\pi_{1/2}$ & ReinMax sampling probabilities; one-hot sample; softmax Jacobian; ReinMax midpoint \\
$a_t$, $\hat a_t$, $w_o$, $H_G^{\mathrm{res}}$ & \texttt{bf16} and quantized residual-stream vectors; channel weights; output factor at the residual stream \\
\multicolumn{2}{l}{\emph{Stages 3--4: recalibration and QAT (\cref{sec:stage2,sec:stage4})}} \\
$r$, $\psi$, $\nu$ & log-ratio of a scale; offset of an RMSNorm gain; the gain \\
$w_r$, $w_\psi$, $w_h$ & weights of the scale and gain penalties and of the hidden-state term \\
$u_{ij}$, $\bar u_{ij}$, $\alpha$, $\beta$ & QAT latent; latent clipped to the grid; initialization fraction; clip margin ($<\tfrac12$) \\
$h_{k,t}$, $\mathcal{S}$ & output of transformer block $k$ at token $t$; blocks used for hidden-state distillation \\
\multicolumn{2}{l}{\emph{Stage 5: selection (\cref{sec:stage5})}} \\
$P$, $\mathrm{acc}^A_i$, $\Delta_i$, $\chi_i$ & number of problems; accuracy of candidate $A$ on problem $i$; paired difference; random sign \\
\bottomrule
\end{tabular}
\caption{Notation.}
\label{tab:notation}
\end{table}

\section{Evaluation settings}
\label{app:protocol}
Every checkpoint, including the GGUF baselines, is exported to dense \texttt{bf16} and scored by the same lm-eval and vLLM stack with the same settings (\cref{tab:evalsettings}). Thinking-mode results use the models' chat template; all other suites use none.
\begin{table}[H]
\centering
\footnotesize
\setlength{\tabcolsep}{2pt}
\begin{tabular}{L{1.55cm}L{4.6cm}L{3.0cm}L{2.4cm}}
\toprule
Suite & Tasks & Shots / format & Decoding \\
\midrule
7-task core & ARC-e/c, WinoGrande, HellaSwag, PIQA, LAMBADA, MMLU & lm-eval defaults (0-shot), log-likelihood & -- \\
Generative & GSM8K (strict match) & 5-shot, no chat template & greedy \\
 & HumanEval & 0-shot, pass@1 & greedy \\
Tool use & BFCL v4: non-live (syntax-tree match), live, multi-turn, irrelevance & official harness, native tool format & greedy, vLLM server \\
Broad suite & MMLU-Pro, TriviaQA, TruthfulQA, SciQ, OpenBookQA, CommonsenseQA, BoolQ, RACE, ARC-C, BBH-CoT, MATH-500, GSM8K-CoT, IFEval, MBPP, HumanEval+ & task defaults; ARC-C 25-shot; BBH 3-shot CoT; GSM8K-CoT 8-shot & greedy \\
Thinking mode & AIME24, AIME25 (30 problems each), MATH-500 & chat template, thinking on, 0-shot & greedy, 1 sample; and sampled (temperature 0.6, top-$p$ 0.95, top-$k$ 20, 4 samples, strict scoring); 32k new tokens \\
\bottomrule
\end{tabular}
\caption{Evaluation settings. GGUF files are dequantized to \texttt{bf16}; RACE is excluded from retention means because its passages appear in MMLU's auxiliary data.}
\label{tab:evalsettings}
\end{table}

\section{Per-task likelihood scores}
\label{app:pertask}
Each stage raises all seven tasks (\cref{tab:pertask}). After QAT the largest remaining gaps to \texttt{bf16} are on HellaSwag (7.8 points) and MMLU (5.6).
\begin{table}[H]
\centering
\small
\setlength{\tabcolsep}{3pt}
\begin{tabular}{lcccccccc}
\toprule
Qwen3-1.7B & ARC-e & ARC-c & Wino & HSwag & PIQA & LAMB & MMLU & Mean \\
\midrule
\texttt{bf16} & 69.7 & 43.0 & 60.9 & 60.4 & 72.2 & 50.5 & 55.5 & 58.9 \\
\ck{GSQ-hbf16} & 51.4 & 29.9 & 54.3 & 41.5 & 64.9 & 28.4 & 36.4 & 43.8 \\
\rowcolor{ourrow}\ck{Rec-hbf16} & 62.4 & 36.6 & 57.1 & 49.2 & 68.2 & 42.7 & 44.4 & 51.5 \\
\rowcolor{ourrow}\ck{QAT-hbf16} & 66.4 & 38.2 & 59.2 & 52.6 & 69.0 & 47.1 & 49.9 & 54.6 \\
\rowcolor{ourrow}\ck{QAT-h2} & 65.1 & 37.5 & 61.0 & 51.4 & 69.2 & 45.5 & 48.0 & 54.0 \\
\rowcolor{ourrow}\ck{QAT+R-h2} (selected) & 66.3 & 38.6 & 58.3 & 51.2 & 70.4 & 44.7 & 47.0 & 53.8 \\
GGUF Q2\_K~\citep{unsloth-qwen3-1.7b-q2k} & 57.7 & 34.6 & 56.5 & 52.2 & 67.4 & 35.9 & 40.0 & 49.2 \\
\bottomrule
\end{tabular}
\caption{The 7-task means of \cref{tab:stages17} task by task (length-normalized accuracy for ARC, HellaSwag and PIQA). Chance-normalized retention after GSQ / recalibration / QAT: 47.6 / 73.1 / 84.2\%.}
\label{tab:pertask}
\end{table}

\section{Codeword search across depth and the query--key rule}
\label{sec:alloc}
\begin{figure}[H]
\centering
\includegraphics[width=0.6\linewidth]{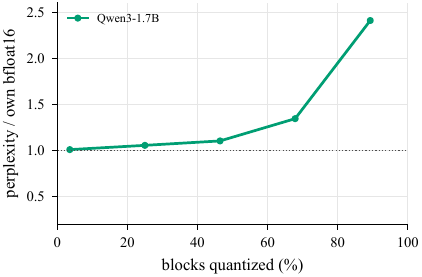}
\caption{GSQ perplexity relative to the model's own \texttt{bf16} as blocks are quantized (before recalibration), Qwen3-1.7B.}
\label{fig:gsq}
\end{figure}

\Cref{fig:gsq} tracks WikiText-2 perplexity as GSQ quantizes one block after another. On Qwen3-1.7B it stays near \texttt{bf16} for the first half of the model and rises over the last third.

\paragraph{Forcing query and key projections to two bits.} \texttt{ThinkBit} keeps query and key projections at $\ge$2 bits, a rule chosen on a model without query--key normalization. Without the rule, the allocator makes 15 of 28 query and 3 of 28 key projections ternary and spends the freed bits on 3-bit layers, at the same 2.0 decoder bits per weight. Every subsequent measurement is close to that of the forced run, and neither reasoning difference is significant (\cref{tab:qk}); on Qwen3-1.7B the rule is unnecessary.
\begin{table}[H]
\centering
\small
\begin{tabular}{lcc}
\toprule
Qwen3-1.7B & query/key at $\ge$2 bits (\texttt{ThinkBit}) & not forced \\
\midrule
ternary query / key projections (of 28) & 0 / 0 & 15 / 3 \\
KronQ objective (relative) & 1.000 & 0.997 \\
WikiText-2 after codeword search & 46.91 & 46.35 \\
held-out KL after recalibration & 0.282 & 0.279 \\
WikiText-2 after recalibration & 19.46 & 19.56 \\
sampled MATH-500 after recalibration & 47.5 & 49.3 \\
\bottomrule
\end{tabular}
\caption{Keeping query and key projections at 2 bits or more changes reasoning by less than 2 points (MATH-500 +1.8, $p{=}0.12$). Otherwise identical settings.}
\label{tab:qk}
\end{table}

\section{Further ablations}
\label{app:morecontrols}
This section details the controls of \cref{sec:controls}. \Cref{tab:starts} compares codeword search with stronger GPTQ starts, \cref{tab:extra} collects the remaining paired contrasts, and \cref{tab:tost} tests equivalence on the 7-task suite. Unless noted, reasoning is sampled, strictly scored MATH-500 with paired tests.

\paragraph{Rounding-only alternatives to QAT.} Moving codewords toward their \texttt{bf16} values without training, then briefly recalibrating the scales, is worse than not moving them (``re-round'' rows of \cref{tab:controls}). GSQ's non-nearest codewords therefore beat nearest rounding; these short controls do not separate QAT's choice of codewords from its longer optimization.

\paragraph{Scales, gains and reverted reassignments.} Recalibrating the scales alone matches recalibrating scales and gains, while the gains alone recover much less (\cref{tab:controls,tab:recmech}). Reverting QAT's MLP reassignments lowers MMLU but raises HumanEval, and reverting its attention reassignments lowers GSM8K (\cref{tab:controls}). These are interventions on a single checkpoint.

\paragraph{Initialization and the reassigned codewords.} Latents that start on their codewords ($\alpha{=}0$) are almost never reassigned, at either learning rate (\cref{app:factorial}). With $\alpha{=}0.5$ the reassignments are single grid steps toward the \texttt{bf16} weight, concentrated in value projections and the 3-bit tier (\cref{app:moves}); their direction largely follows from \cref{eq:residual}, which starts each latent on the side of its codeword where the \texttt{bf16} weight lies.

\paragraph{Rules with a likelihood conjunct.} Two likelihood rules, written before the reasoning results, would have removed stages that the paired reasoning tests show are needed. Codeword search was to stay only if the layer-wise start lost at least 3 MMLU \emph{and} 5 thinking-math points; the MMLU gap is 1.2, so the rule drops the search. Recalibration was to stay if skipping it cost more than one MMLU point; it costs 1.03, a boundary case.

\paragraph{Stronger starts.} To test whether codeword search only beats a weak start, two starts replace the layer-wise one with the same tier map: GPTQ on every tier with MSE-clipped scales, and the same followed by Hessian-weighted Gumbel rounding. Each then receives the same recalibration and QAT. After recalibration their perplexities are close to ours, but their reasoning is more than 20 points lower; QAT narrows the gap without closing it, while the likelihood and code columns barely separate the three (\cref{tab:starts}; paired tests in \cref{tab:reasoning}).
\begin{table}[H]
\centering
\small
\setlength{\tabcolsep}{3.5pt}
\resizebox{\linewidth}{!}{%
\begin{tabular}{lccccccccc}
\toprule
 & \multicolumn{2}{c}{WikiText-2} & held-out & \multicolumn{2}{c}{MATH-500} & \multicolumn{4}{c}{after QAT} \\
\cmidrule(lr){2-3}\cmidrule(lr){5-6}\cmidrule(l){7-10}
Qwen3-1.7B start & before rec. & after rec. & KL & after rec. & after QAT & 7-task & GSM8K & HumanEval & BFCL NL \\
\midrule
\rowcolor{ourrow}GSQ (framework) & \textbf{46.91} & \textbf{19.46} & \textbf{0.282} & \textbf{47.5} & \textbf{43.0} & \textbf{54.6} & 50.2 & 17.7 & \textbf{75.5} \\
GPTQ, all tiers, clipped scales & 1,379 & 22.02 & 0.402 & 18.5 & 32.6 & 53.8 & 49.4 & 17.7 & 70.0 \\
GPTQ + Hessian-weighted rounding & 10,352 & 21.06 & 0.367 & 24.1 & 37.8 & 53.8 & \textbf{51.9} & 17.7 & 71.9 \\
\bottomrule
\end{tabular}}
\caption{ Codeword search beats stronger GPTQ starts on reasoning after the same recalibration (rec.) and QAT, while likelihood barely separates them. Bold: best per column.}
\label{tab:starts}
\end{table}

\paragraph{Remaining paired contrasts.} \Cref{tab:extra} lists contrasts on the number of recalibration passes, on recalibration at 0.52\,GB and on head precision. A second recalibration pass that quantizes the head and embedding beats a single pass on MATH-500 but not on the selector; with either, the recalibrated checkpoint beats the QAT checkpoint at 0.52\,GB. Recalibration before QAT still helps at 0.52\,GB, also under budget forcing, but misses the 4-point threshold set in advance, so the 0.98\,GB estimate of \cref{tab:reasoning} is not claimed at this size. Quantizing the head and embedding of the 14B recalibrated checkpoint doesn't change anything significantly.
\begin{table}[H]
\centering
\small
\setlength{\tabcolsep}{3.5pt}
\resizebox{\linewidth}{!}{%
\begin{tabular}{llccccccc}
\toprule
 & & \multicolumn{4}{c}{sampled MATH-500} & \multicolumn{2}{c}{selector} \\
\cmidrule(lr){3-6}\cmidrule(l){7-8}
Model & Contrast (A vs.\ B) & A & B & $\Delta$ [95\% CI] & $p$ & $\Delta$ [95\% CI] & $p$ \\
\midrule
Qwen3-1.7B & \ck{Rec-h2}: two recalibration passes vs.\ one & 46.7 & 44.4 & +2.3 [0.3, 4.2] & 0.027 & +0.6 & 0.71 \\
 & \ck{Rec-h2} (1 pass) vs.\ \ck{QAT-h2} & 44.4 & 42.0 & -- & -- & +5.1 [2.4, 8.0] & $<10^{-3}$ \\
 & \ck{Rec-hbf16}: two passes vs.\ one pass & 46.6 & 47.5 & $-$0.9 & 0.38 & $-$2.3 & 0.052 \\
 & two passes: \ck{Rec-h2} vs.\ \ck{Rec-hbf16} & 46.7 & 46.6 & +0.1 & 0.96 & -- & -- \\
\rowcolor{ourrow} & \ck{QAT-h2} vs.\ \ck{QAT-noRec-h2} & 42.0 & 38.2 & +3.8 [1.7, 6.0] & $<10^{-3}$ & +3.2 [1.0, 5.6] & 0.009 \\
\midrule
Qwen3-14B & \ck{Rec-h2} vs.\ \ck{Rec-hbf16} & 92.7 & 92.8 & $-$0.15 [$-$1.45, 1.20] & 0.88 & $-$0.4 [$-$2.2, 1.4] & 0.73 \\
\bottomrule
\end{tabular}}
\caption{ Quantizing the head and embedding costs no detectable accuracy. Further paired contrasts (pointwise tests, not Holm-corrected).}
\label{tab:extra}
\end{table}

\paragraph{Same-data QAT baseline.} Our application of EfficientQAT~\citep{chen2024efficientqat} to Qwen3 trains blocks and then the quantization parameters end to end, with 2-bit codewords, per-group zero points and 16-bit scales (groups of 128, 2.14 bits per decoder weight; \texttt{bf16} head and embedding). Trained on our data for the same 805M tokens as recalibration plus QAT, it is close to our checkpoints on likelihood but trails them on reasoning, mostly because fewer of its traces close (\cref{tab:decompall}).

\paragraph{Training data.} Removing Wikipedia and MMLU-auxiliary data, and then Evol-Instruct code, from QAT changes the 7-task mean by at most 0.4 (\cref{tab:controls}); these ablations do not cover earlier stages or the self-generated replies.

\paragraph{Equivalence tests on likelihood.} The 7-task suite was re-run on eight Qwen3-1.7B checkpoints with per-item logging. For each contrast with the QAT checkpoint, a paired bootstrap gives a 90\% interval, and two one-sided tests at level 0.05 declare equivalence when it lies inside a $\pm$1.5-point margin fixed in advance (\cref{tab:tost}). Every removal is equivalent on the 7-task mean. MMLU is not equivalent for four removals, but these differences are far smaller than the MATH-500 differences of the same contrasts. QAT against the recalibrated checkpoint is equivalent on neither measure.
\begin{table}[H]
\centering
\small
\setlength{\tabcolsep}{4pt}
\begin{tabular}{lcccc}
\toprule
QAT vs. & 7-task $\Delta$ [90\% CI] & equiv. & MMLU $\Delta$ [90\% CI] & equiv. \\
\midrule
\ck{LW-QAT}: layer-wise start instead of GSQ & +0.91 [0.37, 1.44] & yes & +1.22 [0.61, 1.84] & no \\
without the scale recalibration & +0.18 [-0.29, 0.63] & yes & +1.03 [0.47, 1.59] & no \\
\ck{QAT-only} (neither) & -0.15 [-0.64, 0.34] & yes & -0.02 [-0.69, 0.65] & yes \\
uniform 2-bit, GPTQ, recalibration & +0.20 [-0.35, 0.73] & yes & +0.72 [0.08, 1.39] & yes \\
tuned GPTQ start & +0.79 [0.32, 1.28] & yes & +2.07 [1.47, 2.68] & no \\
GPTQ + Hessian-weighted rounding & +0.80 [0.31, 1.32] & yes & +1.62 [0.98, 2.29] & no \\
\ck{Rec} (no QAT) & +3.13 [2.63, 3.61] & no & +5.54 [4.91, 6.19] & no \\
\bottomrule
\end{tabular}
\caption{ Every stage removal is equivalent to the full framework on the 7-task mean (margin $\pm$1.5); skipping QAT is not. $\Delta$: \ck{QAT-hbf16} minus the named checkpoint.}
\label{tab:tost}
\end{table}

\section{Paired tests by size, head precision and scale}
\label{app:filesize}
\Cref{tab:filesize} extends the paired tests of \cref{tab:reasoning}. At 0.52\,GB the recalibrated checkpoint beats both the QAT checkpoint and the larger Q2\_K file~\citep{unsloth-qwen3-1.7b-q2k}; quantizing the head and embedding costs no detectable accuracy; and at 14B every two-bit checkpoint beats the smallest GGUF~\citep{unsloth-qwen3-14b-udiq2xxs}, while the pre-registered QAT-versus-QAT-only contrast is not significant.
\begin{table}[H]
\centering
\small
\setlength{\tabcolsep}{4pt}
\resizebox{\linewidth}{!}{%
\begin{tabular}{llccccc}
\toprule
Model & Contrast (A vs.\ B) & A & B & $\Delta$ [95\% CI] & $p$ & status \\
\midrule
Qwen3-1.7B & \ck{QAT-h2} vs.\ \ck{Rec-h2} (0.52\,GB) & 42.0 & 46.7 & \textbf{$-$4.7} [$-$7.1, $-$2.2] & $4\cdot10^{-4}$ & post hoc \\
\rowcolor{ourrow} & \ck{Rec-h2} vs.\ GGUF Q2\_K~\citep{unsloth-qwen3-1.7b-q2k} (0.78\,GB) & 46.7 & 41.9 & \textbf{+4.8} [2.0, 7.7] & 0.001 & post hoc \\
 & \ck{Rec-h2} vs.\ \ck{Rec-hbf16} (0.52 vs.\ 0.98\,GB) & 46.7 & 47.5 & $-$0.8 [$-$2.7, 1.0] & 0.44 & post hoc \\
\midrule
Qwen3-14B & \ck{Rec-hbf16} (6.4\,GB) vs.\ \ck{QAT-h2} (3.74\,GB) & 92.8 & 90.5 & \textbf{+2.3} [0.9, 3.7] & 0.002 & post hoc \\
 & \ck{QAT-h2} vs.\ \ck{Rec-h2} (3.74\,GB) & 90.5 & 92.7 & $-$2.2 [$-$3.8, $-$0.6] & 0.011 & post hoc \\
 & codeword search at the recalibrated checkpoint (\ck{Rec-hbf16} vs.\ \ck{LW-Rec-hbf16}; exploratory) & 92.8 & 89.7 & \textbf{+3.2} [1.8, 4.7] & $10^{-4}$ & post hoc \\
 & \quad same, AIME@16$^\S$ & 48.2 & 36.7 & \textbf{+11.6} [7.3, 16.1] & $<10^{-4}$ & post hoc \\
 & \ck{QAT-only-h2} (3.74\,GB) vs.\ \ck{LW-Rec-hbf16} (6.4\,GB) & 89.6 & 89.7 & $-$0.1 [$-$1.7, 1.5] & 1.0 & post hoc \\
 & \ck{QAT-only-h2} (3.74\,GB) vs.\ GGUF UD-IQ2\_XXS~\citep{unsloth-qwen3-14b-udiq2xxs} (4.47\,GB) & 89.6 & 85.2 & \textbf{+4.4} [2.6, 6.2] & $10^{-4}$ & post hoc \\
 & \ck{QAT-h2} vs.\ \ck{QAT-only-h2} (pre-registered) & 90.5 & 89.6 & +0.9 [$-$0.4, 2.3] & 0.18 & pre-reg. \\
 & \quad same, AIME@16 (secondary) & 40.9 & 37.2 & +3.8 [0.7, 7.0] & 0.032 & pre-reg. \\
 & QAT (3.74\,GB) vs.\ GGUF UD-IQ2\_XXS~\citep{unsloth-qwen3-14b-udiq2xxs} (4.47\,GB) & 90.5 & 85.2 & \textbf{+5.3} [3.5, 7.1] & $<10^{-4}$ & post hoc \\
\rowcolor{ourrow} & \ck{Rec-h2} (base-mixture selection, 3.74\,GB) vs.\ GGUF UD-IQ2\_XXS~\citep{unsloth-qwen3-14b-udiq2xxs} (4.47\,GB) & 92.7 & 85.2 & \textbf{+7.5} [5.7, 9.5] & $<10^{-4}$ & check \\
\bottomrule
\end{tabular}}
\caption{ At 0.52\,GB, \ck{Rec-h2} beats \ck{QAT-h2} and the larger Q2\_K; at 14B every two-bit checkpoint beats the smallest GGUF. Paired MATH-500 contrasts (AIME@16 where noted). Bold: significant after Holm correction together with \cref{tab:reasoning}. $^\S$Separate evaluation run. Status: pre-reg., pre-registered; post hoc, chosen after earlier results; check, robustness check.}
\label{tab:filesize}
\end{table}

\section{How checkpoints are selected}
\label{app:selaudit}
The selector samples four answers per problem on 313 level-3--5 MATH training problems that are disjoint from MATH-500 and from the training prompts, scores them strictly, and keeps the higher mean when the paired sign-flip test gives $p\le0.05$, otherwise the smaller checkpoint. \Cref{tab:selaudit} lists every decision; MATH-500 agrees with each of them.
\begin{table}[H]
\centering
\small
\setlength{\tabcolsep}{3pt}
\resizebox{0.6\linewidth}{!}{%
\begin{tabular}{llcccl}
\toprule
Model & A vs.\ B & A & B & $p$ & kept \\
\midrule
Qwen3-1.7B & \ck{QAT-hbf16} vs.\ \ck{Rec-hbf16} & 22.5 & 28.8 & $<10^{-3}$ & \ck{Rec-hbf16} \\
\rowcolor{ourrow}Qwen3-1.7B & \ck{QAT-h2} vs.\ \ck{Rec-h2} & 21.9 & 27.6 & $<10^{-3}$ & \ck{Rec-h2} \\
Qwen3-1.7B & \ck{Rec-hbf16} vs.\ \ck{Rec-h2} & 28.8 & 27.6 & 0.37 & \ck{Rec-h2} \\
Qwen3-14B & \ck{QAT-h2} vs.\ \ck{Rec-hbf16} & 83.3 & 87.9 & $<10^{-3}$ & \ck{Rec-hbf16} \\
\rowcolor{ourrow}Qwen3-14B & \ck{QAT-h2} vs.\ \ck{Rec-h2} & 83.3 & 87.5 & $<10^{-3}$ & \ck{Rec-h2} \\
Qwen3-14B & \ck{Rec-hbf16} vs.\ \ck{Rec-h2} & 87.9 & 87.5 & 0.73 & \ck{Rec-h2} \\
Qwen3-8B & \ck{Rec-hbf16} vs.\ \ck{Rec-h2} & 75.6 & 76.1 & 0.72 & \ck{Rec-h2} \\
\bottomrule
\end{tabular}}
\caption{ Under the base mixture the selector kept the recalibrated checkpoint in every decision, and MATH-500 agrees. A, B: strict accuracy on the 313 selection problems.}
\label{tab:selaudit}
\end{table}

\section{What the recalibration changes}
\label{app:recmech}
With codewords frozen, training the scales alone lowers held-out KL from 0.641 to 0.282, the same as training scales and RMSNorm gains together, while the gains alone reach 0.418 (\cref{tab:recmech}).
\begin{table}[H]
\centering
\small
\begin{tabular}{lcc}
\toprule
Qwen3-1.7B, codewords frozen & Held-out KL & WikiText-2 \\
\midrule
\ck{GSQ-hbf16} & 0.641 & 46.91 \\
+ RMSNorm gains only & 0.418 & 25.29 \\
+ scales only & 0.282 & 19.49 \\
\rowcolor{ourrow}+ scales and RMSNorm gains (\ck{Rec-hbf16}) & 0.282 & 19.46 \\
\bottomrule
\end{tabular}
\caption{The scales carry recalibration: scales alone reach the same held-out KL as scales plus RMSNorm gains. Qwen3-1.7B, same data and schedule.}
\label{tab:recmech}
\end{table}

\Cref{tab:scalestats} compares three scales for each group $g$ of the frozen codewords $c_g$: the GSQ scale $s_{0,g}$, the recalibrated scale $s_g=s_{0,g}e^{r_g}$ (\cref{eq:recal}), and the weight-space least-squares scale $s^\star_g=\langle W_g,c_g\rangle/\langle c_g,c_g\rangle$. If recalibration repaired weight error, $r_g$ would follow $\log(s^\star_g/s_{0,g})$. It does not: the signs agree in about half of the groups in every tier, the correlation is at most 0.08, and weight error rises in every tier.
\begin{table}[H]
\centering
\small
\setlength{\tabcolsep}{4pt}
\begin{tabular}{lcccccc}
\toprule
 & median & same sign & & \multicolumn{3}{c}{relative weight error} \\
\cmidrule(l){5-7}
Tier & $|r|$ & as $s^\star$ (\%) & corr. & GSQ $s_0$ & recalibrated $s$ & least-squares $s^\star$ \\
\midrule
ternary & 0.064 & 51.6 & 0.08 & 0.425 & 0.431 & 0.325 \\
2-bit & 0.054 & 49.8 & 0.00 & 0.198 & 0.207 & 0.192 \\
3-bit & 0.058 & 50.8 & 0.05 & 0.287 & 0.296 & 0.109 \\
\rowcolor{ourrow}all & 0.062 & 51.4 & 0.07 & 0.386 & 0.393 & 0.277 \\
\bottomrule
\end{tabular}
\caption{ Recalibration does not move scales toward the weight-space optimum, and weight error rises in every tier. Per-group scale changes on Qwen3-1.7B; corr.: correlation of $r$ with $\log(s^\star/s_0)$.}
\label{tab:scalestats}
\end{table}

\Cref{tab:blockerr} measures where the output improves. The local error feeds each quantized block the \texttt{bf16} input of that block; the propagated error runs the whole quantized model. Recalibration and QAT raise both errors, averaged over blocks, while lowering the KL of the next-token distribution.
\begin{table}[H]
\centering
\small
\begin{tabular}{lccc}
\toprule
Qwen3-1.7B & local block error & propagated error & next-token KL \\
\midrule
\ck{GSQ-hbf16} & 0.079 & 0.214 & 1.300 \\
\rowcolor{ourrow}\ck{Rec-hbf16} & 0.094 & 0.282 & 0.571 \\
\rowcolor{ourrow}\ck{QAT-hbf16} & 0.122 & 0.373 & 0.371 \\
\bottomrule
\end{tabular}
\caption{ Recalibration and QAT make single blocks less faithful but bring the output distribution closer to \texttt{bf16}. Relative block-output error (local: \texttt{bf16} input; propagated: fully quantized model) and output KL on WikiText-2.}
\label{tab:blockerr}
\end{table}

\section{QAT initialization and learning rate}
\label{app:factorial}
\begin{figure}[H]
\centering
\includegraphics[width=0.75\linewidth]{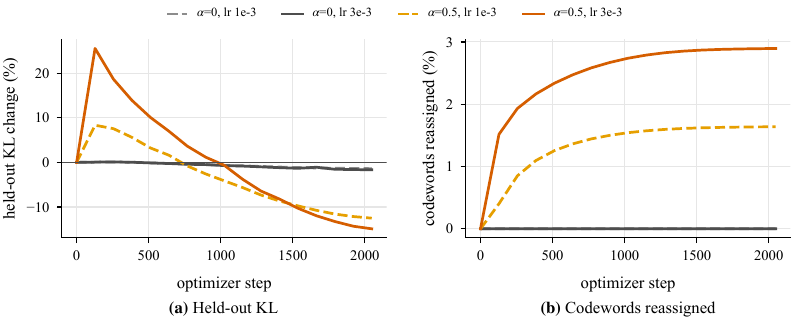}
\caption{QAT latent initialization $\alpha$ $\times$ learning rate on Qwen3-1.7B (short runs from the recalibrated model). \textbf{(a)}~Held-out KL relative to the start. \textbf{(b)}~Share of codewords changed. With $\alpha=0$ fewer than 0.003\% of codewords are reassigned. Validation sequences are the chunks that follow each run's training slice, so runs with different training lengths are scored on different held-out text; this is why the starting KL (0.251) differs from long QAT's (0.311). KL values are comparable within a run, not across runs.}
\label{fig:factorial}
\end{figure}

\Cref{fig:factorial} crosses the latent initialization $\alpha$ of \cref{eq:residual} with the learning rate. Latents that start on their codewords ($\alpha{=}0$) almost never cross a rounding boundary at either learning rate. With $\alpha{=}0.5$, KL first rises while codewords are reassigned and then falls below its starting value.

\section{Where QAT reassigns codewords}
\label{app:moves}
\begin{figure}[H]
\centering
\includegraphics[width=\linewidth]{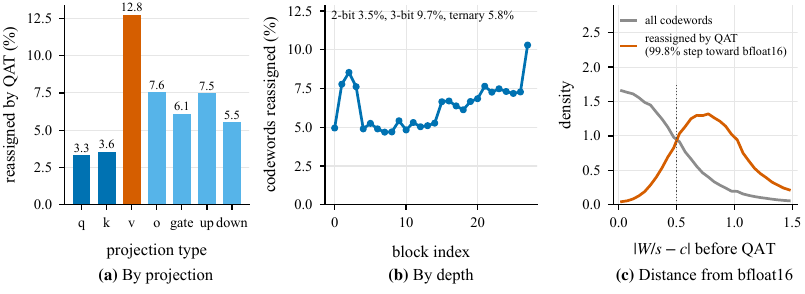}
\caption{Where long QAT reassigns codewords on Qwen3-1.7B. \textbf{(a)}~Share of each projection type's codewords that QAT changes; value projections (orange) change most, query and key projections (dark blue) least. \textbf{(b)}~Share changed in each transformer block, with the share per tier in the inset text. Codewords at the first and last transformer blocks are reassigned more than other blocks\textbf{(c)}~Distribution, before QAT, of the distance $|W/s-c|$ between each codeword $c$ and its \texttt{bf16} weight $W/s$, in grid steps with the recalibrated scales, for all codewords (grey) and for those that QAT reassigns (orange). The dotted line at 0.5 is the rounding boundary: most reassigned codewords lie beyond it, and 99.8\% of reassignments move one step toward the \texttt{bf16} weight.}
\label{fig:moves}
\end{figure}

QAT reassigns 6.4\% of the codewords. \Cref{fig:moves} shows that the reassignments concentrate in value projections, the 3-bit tier and the last block, and that the reassigned codewords are mostly those already more than half a grid step from their \texttt{bf16} value; almost every reassignment is a single step toward that value.

\section{Forcing an answer at a token budget}
\label{app:budget}
Budget forcing~\citep{muennighoff2025s1} stops the thinking after a budget of 1k--16k tokens and asks for an answer, which separates finishing a trace from solving the problem. With a forced answer the 0.52\,GB checkpoints tie Q2\_K~\citep{unsloth-qwen3-1.7b-q2k} at every budget, so their lead without forcing comes from finishing more traces (\cref{tab:budget}).
\begin{table}[H]
\centering
\small
\begin{tabular}{lccccc|c}
\toprule
Thinking budget (tokens) & 1k & 2k & 4k & 8k & 16k & 32k, unforced \\
\midrule
\texttt{bf16} & 74.2 & 77.0 & 82.4 & 86.0 & 90.0 & 90.2 \\
\rowcolor{ourrow}\ck{Rec-hbf16} & 40.4 & 43.2 & 44.4 & 44.6 & 45.2 & 40.6 \\
\ck{QAT-hbf16} & 42.2 & 43.6 & 44.4 & 43.4 & 39.2 & 39.0 \\
\ck{QAT-only-hbf16} & 33.6 & 34.6 & 36.6 & 31.0 & 28.8 & 26.6 \\
\midrule
\multicolumn{7}{l}{\emph{0.52\,GB checkpoints (2-bit head and embedding) and Q2\_K}} \\
\rowcolor{ourrow}\ck{Rec-h2}, 0.52\,GB & 40.6 & 42.0 & 43.4 & 42.6 & 41.0 & 39.0 \\
\ck{QAT-h2}, 0.52\,GB & 40.2 & 43.6 & 45.8 & 42.6 & 37.2 & 38.8 \\
GGUF Q2\_K~\citep{unsloth-qwen3-1.7b-q2k}, 0.78\,GB & 41.4 & 41.8 & 43.2 & 40.4 & 41.4 & 27.8 \\
GGUF UD-IQ2\_M~\citep{unsloth-qwen3-1.7b-udiq2m}, 0.71\,GB & 28.2 & 26.8 & 27.8 & 25.6 & 24.8 & 14.2 \\
GGUF UD-IQ2\_XXS~\citep{unsloth-qwen3-1.7b-udiq2xxs}, 0.61\,GB & 4.0 & 4.0 & 4.6 & 4.0 & 3.8 & 2.8 \\
\bottomrule
\end{tabular}
\caption{ With a forced answer the 0.52\,GB checkpoints and Q2\_K tie, so the strict lead of \ck{Rec-h2} comes from finishing more traces. Greedy thinking-mode MATH-500 on Qwen3-1.7B with the answer forced after the given number of thinking tokens.}
\label{tab:budget}
\end{table}

\section{Trace closing and accuracy on closed traces}
\label{app:matched}
\Cref{tab:matched} splits each paired MATH-500 contrast into a change in how often traces close and a change in accuracy on closed traces. Codeword search mainly improves accuracy, whereas the recalibration and QAT mainly change how often traces close.
\begin{table}[H]
\centering
\small
\begin{tabular}{llccc}
\toprule
Model & Contrast & $\Delta$ closing rate & $\Delta$ accuracy on closed & problems \\
\midrule
\rowcolor{ourrow}Qwen3-1.7B & codeword search & +1.4 [$-$1.0, 3.8] & \textbf{+10.4} [7.1, 13.6] & 414 \\
 & scale recalibration & \textbf{+12.3} [9.8, 14.9] & +2.1 [$-$1.4, 5.8] & 379 \\
 & \ck{QAT} vs.\ \ck{Rec} & \textbf{$-$18.4} [$-$21.1, $-$15.6] & $-$0.5 [$-$3.5, 2.5] & 440 \\
\bottomrule
\end{tabular}
\caption{Codeword search mainly raises accuracy on closed traces; recalibration and QAT mainly change how many traces close. Problem-matched sampled MATH-500 differences (95\% CI); bold: $p{<}0.01$.}
\label{tab:matched}
\end{table}

\section{Packed files and serving}
\label{app:packed}
Training checkpoints store every codeword in a 4-bit container. We re-pack checkpoints at their true widths and decode them again; every codeword round-trips bit-exactly and lies on its layer's grid, and a reference decoder that reads only the packed file rebuilds a checkpoint that loads in stock \texttt{transformers} (\cref{tab:packed}). The measured sizes are within 0.6\% of those computed from the grids. For Qwen3-1.7B the \texttt{bf16} tied embedding is about 64\% of the 0.977\,GB file, so quantizing the head and embedding matters at small scale.
\begin{table}[H]
\centering
\small
\begin{tabular}{llcccc}
\toprule
 & & quantized & \multicolumn{2}{c}{bits per quantized weight} & \\
\cmidrule(lr){4-5}
Model & Checkpoint & weights & measured & computed & file (GB) \\
\midrule
Qwen3-1.7B & \ck{QAT-hbf16} & 1.41B & 2.011 & 2.000 & 0.977 \\
\rowcolor{ourrow} & \ck{QAT-h2} & 2.03B & 2.046 & 2.038 & 0.520 \\
\rowcolor{ourrow} & \ck{Rec-h2} & 2.03B & 2.046 & 2.038 & 0.520 \\
\midrule
Qwen3-14B & \ck{QAT-only-h2} & 14.77B & 2.023 & 2.013 & 3.736 \\
\rowcolor{ourrow} & \ck{Rec-h2} & 14.77B & 2.023 & 2.013 & 3.736 \\
\midrule
\rowcolor{ourrow}Qwen3-8B & \ck{Rec-h2} & 8.19B & 2.029 & 2.019 & 2.078 \\
\bottomrule
\end{tabular}
\caption{Packed files decode bit-exactly and are within 0.6\% of the sizes computed from the grids. Bits per quantized weight include the group scales.}
\label{tab:packed}
\end{table}
vLLM serves our checkpoints directly from their 4-bit containers (compressed-tensors format, Machete kernels; decoder weights are never expanded, embedding and head stay in \texttt{bf16}) on one H100 (\cref{tab:serve}). These kernels read 4 bits per codeword, so memory and speed reflect the container rather than the 2-bit widths of the packed file; at 1.7B the \texttt{bf16} embedding and head dominate and decoding is not limited by weight bandwidth.
\begin{table}[H]
\centering
\small
\setlength{\tabcolsep}{4pt}
\begin{tabular}{lccccc}
\toprule
 & weight memory & \multicolumn{2}{c}{decode (tokens/s)} & prefill, 1{,}024 & top-1 agreement \\
\cmidrule(lr){3-4}
Qwen3 & (GiB) & batch 1 & batch 16 & tokens (ms) & (\%) \\
\midrule
1.7B & 3.22 $\to$ 1.85 & 424 $\to$ 408 & 6{,}232 $\to$ 5{,}998 & 12.8 $\to$ 15.7 & 98.7 \\
8B & 15.27 $\to$ 5.68 & 146 $\to$ 212 & 2{,}225 $\to$ 3{,}184 & 27.0 $\to$ 28.4 & 99.1 \\
14B & 27.52 $\to$ 9.31 & 86 $\to$ 150 & 1{,}341 $\to$ 2{,}302 & 48.0 $\to$ 45.0 & 98.3 \\
\bottomrule
\end{tabular}
\caption{Serving from packed containers cuts weight memory 1.7--3.0$\times$ and speeds up batch-1 decoding at 8B and 14B. vLLM on one H100; memory in GiB.}
\label{tab:serve}
\end{table}

\section{Retention on a broad benchmark suite}
\label{app:thorough}
\begin{figure}[H]
\centering
\includegraphics[width=0.85\linewidth]{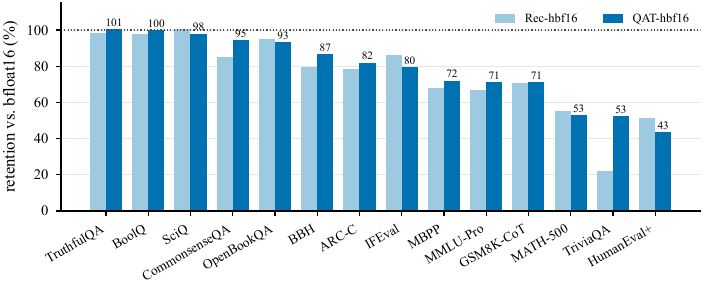}
\caption{Qwen3-1.7B: retention on 14 benchmarks after recalibration and after QAT (\ck{Rec-hbf16}, \ck{QAT-hbf16}) (RACE omitted; its test passages appear in MMLU's auxiliary training data). Numbers above bars are QAT retention.}
\label{fig:thorough}
\end{figure}

Most knowledge and commonsense tasks keep 80\% or more of \texttt{bf16}'s score after QAT, while MATH-500, TriviaQA and HumanEval+ keep about half (\cref{fig:thorough}).

\section{Accuracy by model size in bytes}
\label{app:frontier}
\begin{figure}[H]
\centering
\includegraphics[width=0.8\linewidth]{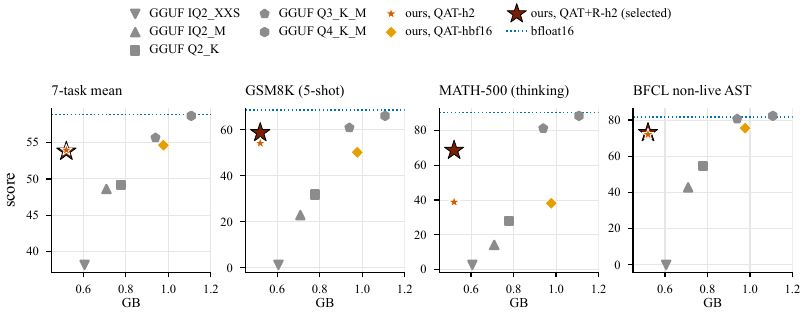}
\caption{Qwen3-1.7B: score against whole-model bytes (generation metrics greedy). Grey: published llama.cpp quantizations (Unsloth~\citep{unsloth2025qwen3gguf}; the UD- prefix is omitted in labels). Red star: ours with a 2-bit head and embedding (\ck{QAT-h2}); orange diamond: ours with the \texttt{bf16} tied embedding (\ck{QAT-hbf16}). Dotted: \texttt{bf16}. Over 30 individually scored sub-scores (RACE excluded; greedy MATH counted; differences of at most 0.5 points count as ties) the 0.52\,GB QAT checkpoint (\ck{QAT-h2}) wins 23--26 against each 2-bit GGUF.}
\label{fig:frontier}
\end{figure}

\Cref{fig:frontier} places the Qwen3-1.7B checkpoints among the GGUF files of the same model on four benchmarks, and \cref{tab:decompall} splits sampled MATH-500 into closing rate and accuracy on closed traces for our checkpoints, the GGUF files of \cref{fig:frontierall} and the EfficientQAT port.

\begin{table}[H]
\centering
\small
\setlength{\tabcolsep}{4pt}
\begin{tabular}{llccccc}
\toprule
Model & Checkpoint & GB & strict & lenient & closed (\%) & strict if closed \\
\midrule
\multicolumn{7}{l}{\emph{$\le$2.6 bits per parameter}} \\
\rowcolor{ourrow}Qwen3-1.7B & \textbf{\ck{QAT+R-h2} (selected)} & 0.52 & 73.0 & 74.6 & 87.1 & 83.8 \\
\rowcolor{ourrow}Qwen3-1.7B & \ck{Rec-h2} & 0.52 & 46.7 & 47.3 & 86.0 & 54.2 \\
\rowcolor{ourrow}Qwen3-1.7B & \ck{QAT-h2} & 0.52 & 42.0 & 44.9 & 70.0 & 60.0 \\
\multicolumn{7}{l}{\emph{2.6--4 bits per parameter}} \\
Qwen3-1.7B & GGUF UD-IQ2\_XXS~\citep{unsloth-qwen3-1.7b-udiq2xxs} & 0.61 & 1.0 & 2.8 & 46.8 & 2.1 \\
Qwen3-1.7B & GGUF UD-IQ2\_M~\citep{unsloth-qwen3-1.7b-udiq2m} & 0.71 & 24.6 & 27.1 & 46.7 & 52.6 \\
Qwen3-1.7B & GGUF Q2\_K~\citep{unsloth-qwen3-1.7b-q2k} & 0.78 & 41.9 & 44.8 & 56.0 & 74.7 \\
\multicolumn{7}{l}{\emph{$>$4 bits per parameter}} \\
Qwen3-1.7B & GGUF Q3\_K\_M~\citep{unsloth-qwen3-1.7b-q3km} & 0.94 & 84.3 & 84.7 & 98.4 & 85.7 \\
Qwen3-1.7B & EfficientQAT port, same data & 1.00 & 28.8 & 32.8 & 55.9 & 51.4 \\
Qwen3-1.7B & GGUF Q4\_K\_M~\citep{unsloth-qwen3-1.7b-q4km} & 1.11 & 90.8 & 90.9 & 99.3 & 91.4 \\
\midrule
\multicolumn{7}{l}{\emph{$\le$2.6 bits per parameter}} \\
\rowcolor{ourrow}Qwen3-8B & \textbf{\ck{QAT+R-h2} (selected)} & 2.08 & 92.2 & 93.2 & 95.4 & 96.7 \\
\rowcolor{ourrow}Qwen3-8B & \ck{Rec-h2} & 2.08 & 86.6 & 87.3 & 94.5 & 91.6 \\
Qwen3-8B & GGUF UD-IQ2\_XXS~\citep{unsloth-qwen3-8b-udiq2xxs} & 2.61 & 35.2 & 49.4 & 39.6 & 89.0 \\
\multicolumn{7}{l}{\emph{2.6--4 bits per parameter}} \\
Qwen3-8B & GGUF UD-IQ2\_M~\citep{unsloth-qwen3-8b-udiq2m} & 3.11 & 92.0 & 92.7 & 96.7 & 95.1 \\
Qwen3-8B & GGUF Q2\_K~\citep{unsloth-qwen3-8b-q2k} & 3.28 & 91.1 & 92.5 & 95.0 & 95.8 \\
\midrule
\multicolumn{7}{l}{\emph{$\le$2.6 bits per parameter}} \\
\rowcolor{ourrow}Qwen3-14B & \textbf{\ck{QAT+R-h2} (selected)} & 3.74 & 94.8 & 95.4 & 98.3 & 96.5 \\
\rowcolor{ourrow}Qwen3-14B & \ck{Rec-h2} & 3.74 & 92.7 & 93.0 & 96.5 & 96.1 \\
\rowcolor{ourrow}Qwen3-14B & \ck{QAT-h2} & 3.74 & 90.5 & 91.2 & 96.8 & 93.6 \\
Qwen3-14B & GGUF UD-IQ2\_XXS~\citep{unsloth-qwen3-14b-udiq2xxs} & 4.47 & 85.2 & 85.8 & 94.6 & 90.1 \\
\multicolumn{7}{l}{\emph{2.6--4 bits per parameter}} \\
Qwen3-14B & GGUF UD-IQ2\_M~\citep{unsloth-qwen3-14b-udiq2m} & 5.43 & 95.0 & 95.2 & 98.9 & 96.0 \\
Qwen3-14B & GGUF Q2\_K~\citep{unsloth-qwen3-14b-q2k} & 5.75 & 95.0 & 95.0 & 98.8 & 96.1 \\
\midrule
\multicolumn{7}{l}{\emph{Smaller model (reference)}} \\
Qwen3-4B & GGUF Q4\_K\_M~\citep{unsloth-qwen3-4b-q4km} & 2.50 & 96.2 & 96.3 & 99.6 & 96.6 \\
\bottomrule
\end{tabular}
\caption{ \ck{QAT+R-h2} closes more traces than Q2\_K and is more accurate on them; the earlier 0.52\,GB checkpoints win only by closing more traces. Sampled MATH-500 split into traces that close within 32k tokens and strict accuracy on them. Rows grouped by bits per parameter.}
\label{tab:decompall}
\end{table}
The 0.52\,GB recalibrated checkpoint selected under the base mixture (\ck{Rec-h2}) also scores BFCL live 59.9 and multi-turn 2.4 (non-live 73.0) and AIME@16 1.0; like the 2-bit GGUFs of this model, it does not solve competition problems.

\section{Qwen3-14B compared by size}
\label{app:scale14}
At 3.74\,GB our Qwen3-14B checkpoints beat the smallest 14B GGUF (4.47\,GB) on every reasoning column, while the larger 2-bit GGUFs and smaller models at four to five bits per parameter score higher (\cref{tab:scale}).
\begin{table}[H]
\centering
\footnotesize
\setlength{\tabcolsep}{2pt}
\resizebox{\linewidth}{!}{%
\begin{tabular}{@{}lccccccccc@{}}
\toprule
Checkpoint & Size (GB) $\downarrow$ & 7-task $\uparrow$ & MMLU $\uparrow$ & GSM8K $\uparrow$ & HE $\uparrow$ & MT $\uparrow$ & M500 $\uparrow$ & A24/25 $\uparrow$ & A@16 $\uparrow$ \\
\midrule
Qwen3-14B \texttt{bf16} & 29.5 & 74.3 & 77.1 & 88.2 & 57.3 & 40.8 & 97.5 & 76.7/60.0 & 73.6 \\
\midrule
\multicolumn{10}{l}{\emph{Smaller models at 4--5 bits per parameter (reference)}} \\
Qwen3-4B GGUF Q4\_K\_M~\citep{unsloth-qwen3-4b-q4km} & 2.50 & 66.7 & 68.3 & 84.8 & 70.1 & 30.0 & 96.3 & 73.3/50.0 & 65.3 \\
Qwen3-8B GGUF Q3\_K\_M~\citep{unsloth-qwen3-8b-q3km} & 4.12 & 69.2 & 71.0 & 86.5 & 61.0 & 31.6 & 95.8 & 56.7/50.0 & 64.1 \\
\midrule
\multicolumn{10}{l}{\emph{Qwen3-14B, $\le$2.6 bits per parameter}} \\
\rowcolor{ourrow}Ours: \ck{Rec-h2} & 3.74 & 69.9 & 69.5 & 87.7 & 65.2 & 32.0 & 92.7 & 50.0/33.3 & 46.5 \\
\rowcolor{ourrow}Ours: \ck{QAT-h2} & 3.74$^\dagger$ & 70.6 & 70.2 & 86.9 & 60.4 & 27.8 & 90.5 & 43.3/26.7 & 40.9 \\
\ck{QAT-only-h2} & 3.74 & 70.2 & 69.9 & 85.7 & 62.8 & 27.2 & 89.6 & 30.0/20.0 & 37.2 \\
GGUF UD-IQ2\_XXS~\citep{unsloth-qwen3-14b-udiq2xxs} & 4.47 & 66.7 & 65.2 & 71.2 & 39.0 & 20.8 & 85.2 & 13.3/13.3 & 26.7 \\
\multicolumn{10}{l}{\emph{Qwen3-14B, 2.6--4 bits per parameter}} \\
\rowcolor{ourrow}Ours: \ck{Rec-hbf16} & 6.4$^\ast$ & 70.3$^\S$ & 70.1$^\S$ & 87.4$^\S$ & 56.1$^\S$ & 31.0 & 92.8 & 43.3/43.3$^\S$ & 48.2$^\S$ \\
\ck{LW-Rec-hbf16} & 6.4$^\ast$ & 69.1 & 67.9 & 85.2 & 55.5 & 27.8 & 89.7 & 33.3/33.3 & 36.7$^\S$ \\
GGUF UD-IQ2\_M~\citep{unsloth-qwen3-14b-udiq2m} & 5.43 & 71.0 & 72.0 & 85.2 & 65.2 & 32.2 & 95.0 & 66.7/60.0 & 60.5 \\
GGUF Q2\_K~\citep{unsloth-qwen3-14b-q2k} & 5.75 & 71.6 & 73.8 & 87.3 & 54.9 & 30.5 & 95.0 & 66.7/60.0 & 63.0 \\
\bottomrule
\end{tabular}}
\caption{At 3.74\,GB our checkpoints beat UD-IQ2\_XXS (4.47\,GB) on reasoning; Q2\_K and larger files remain stronger. Qwen3-14B by whole-model bytes. M500: sampled MATH-500; A24/25: greedy AIME; A@16: mean pass@1 over 16 samples; HE: HumanEval; MT: BFCL multi-turn. Rows grouped by bits per parameter. $^\dagger$Not packed; $^\ast$computed from the grids; $^\S$separate evaluation run.}
\label{tab:scale}
\end{table}

\end{document}